# How many LLM-labelled rows does a decision encoder need, and how wrong can the labels be?

*Learning curves and label-noise sweeps for a fine-tuned non-autoregressive decision model (Laya)*

Mohammed Nayeem Khan
mohammed.nayeem.khan@intel.com
mnknayeem@gmail.com

Aliyas Ahmad Misri
aliyas.ahmad.misri@intel.com

Intel Corporation

### Abstract

A decision encoder such as Laya reads a question with named options (for example, which department should handle a ticket) and a text, and picks an option. It reads both in one forward pass and returns a probability for every option, without writing any text. We ask how many LLM-written training rows such an encoder needs and how wrong those labels may be. We ran 19 fine-tunes of Laya (ModernBERT-large) on nested subsets of a 2,818-row pool and scored them on 1,065 held-out items from the same generator, so every test item is in-distribution and no out-of-family test was run. Accuracy rises from 69.4% before fine-tuning to 74.3% with 5% of the rows, 84.8% with 50% and 86.9% with all rows. The curve flattens, and a saturating fit with bootstrap intervals cannot pin down its plateau. Seed spread at 20% of the rows (s.d. 1.6 points) exceeds some of the steps between sizes. Giving small subsets as many updates as the full pool gains 1.9 points at 5% and 1.9 points at 20% of the rows (one seed; p=0.14 and 0.13), and at 20% that is below the 2.3-point difference between two ordinary seeds. A single run with 10% symmetric label noise shows no detectable change (-0.2 points, 95% interval -1.7 to +1.4); 20% costs 3.4 points and 40% costs 8.7 points. Confusable-pair noise at 20% costs 7.4 points, and under noise the model moves from over-confident to under-confident. An independent labeller disagrees with the gold labels on 3.5% of held-out items, and about half of its choice disagreements (16 of 33, leaving each item out of the partner map) name the partner option that the confusable-pair noise uses. Laya is +0.9 points ahead of a TF-IDF classifier trained on the same rows at full size (the interval includes zero) and +21.1 points ahead at 5% of the rows. Everything is synthetic, LLM-labelled and from one backbone, one trainer and mostly one seed.

**Key findings (every number recomputed from frozen results)**

1. Accuracy: 69.4% before fine-tuning, 74.3% with 143 rows, 79.8% with 564, 84.8% with 1,413, 86.9% with all 2,818 (n=1,065); the last step is inside one run's binomial error (1.0 points).
2. Seed s.d. is 1.6 points at 20% and 0.7 at 50% of the rows. The step from 20% to 35% is therefore within seed noise, and the step from 50% to 75% is larger than two s.d.; steps between single-seed sizes cannot be judged. The fitted plateau interval is [97.8-100.0]% and is not informative.
3. Equal-update control: +1.9 points at 5% and +1.9 points at 20% of the rows (one seed, McNemar p=0.145 and 0.130), with ECE rising from 6.7% to 20.6% and from 5.3% to 13.1%. A second ordinary seed at 20% differs from the first by 2.3 points.
4. Symmetric noise: -0.2 points at 10% (one run, paired 95% interval -1.7 to +1.4, so not detectable but not excluded), -3.4 at 20%, -8.7 at 40%. Confusable-pair noise at 20%: -7.4 points, with 13.9% of decisions changed against a clean-seed churn of 9.5%. Mean confidence falls below accuracy in every noisy run.
5. The independent labeller disagrees on 3.5% of held-out items and the model errs on 39.9% of those against 12.2% elsewhere. Interpolating between single runs, that rate implies -0.1 points (symmetric) and -1.3 points (confusable-pair). With each item left out of the partner map, 16 of its 33 choice disagreements name the confusable-pair partner (18 with it in, 11 with every contested item left out).
6. Laya minus TF-IDF + LR on the same rows: +21.1 points at 5% and +0.9 at 100% (interval -1.5 to +3.4); to reach 82% the encoder needs about 859 rows against 1,451 for TF-IDF.
7. Scope: synthetic, LLM-labelled, one backbone and trainer, mostly one seed; every test item is in-distribution (same generator as the training rows); nothing is shown about human labels.

## 1 Introduction and short answer

A team that adapts a small decision model to its own routing, triage or approval questions needs two numbers before it writes labels or asks a language model to write them: how many labelled rows to collect, and how much it matters if some labels are wrong. The literature on learning curves and noisy labels answers both in general. The numbers depend on the model, the task and the way the labels were made. We found none for a typed-decision encoder trained on LLM-written rows (Section 14).

We study one such encoder, Laya, adapted with its own trainer. The training pool has 2,818 rows written and screened by language models, and the held-out set has 1,065 items from the same generator. We trained on nested subsets from 5% to 100% of the pool and repeated two sizes with a second seed. The smallest sizes were also trained for as many updates as the full pool gets. We also trained the full pool with a share of the labels replaced by wrong ones, either uniformly (symmetric) or by the option that the clean models already confuse with the gold one (confusable-pair). All numbers come from 19 fine-tunes (including the four full-pool models) and a TF-IDF classifier trained on the same rows.

The shipped model scores 69.4% before any fine-tuning. With 143 rows (about 4 per question-option cell) it reaches 74.3%, with 564 rows 79.8%, and with the full 2,818 rows 86.9%. The curve flattens: the last step from 75% to 100% of the rows moves the mean by -0.2 points, inside the binomial error of one run (1.0 points). On the interpolated seed-mean curve, 82% took about 859 rows and 85% about 1,456. Two seeds at 20% of the rows differ enough (s.d. 1.6 points) that the step to 35% is within seed noise. Several single-seed steps cannot be read at all.

A tenth of the labels flipped at random changes accuracy by -0.2 points in one run (paired 95% interval -1.7 to +1.4). That is no detectable cost, but it does not exclude a cost of up to 1.7 points. At 20% the cost is 3.4 points, and at 40% it is 8.7 points. Confusable-pair noise at 20% costs 7.4 points, more than twice as much as symmetric noise at the same rate (one run each). The independent labeller disagrees with the gold labels on 3.5% of the held-out items. Reading the sweeps at that rate gives about 0.1 points lost for symmetric and 1.3 points lost for confusable-pair errors, so a disagreement rate of that size does not appear to be what limits accuracy here.

Laya is +21.1 points ahead of a TF-IDF logistic regression trained on the same rows at 5% of the rows, and +0.9 points ahead at 100% (interval -1.5 to +3.4, so no difference can be claimed there). The encoder's advantage is a small-data advantage, and on the multilingual task the n-gram model ends up ahead.

The claim is narrow: these measurements for typed-decision encoders on LLM-labelled synthetic data, with seed repeats, an equal-update control and a per-task budget, plus a noise sweep set against an observed labeller disagreement. It says nothing about human labels, other backbones or real traffic. Every test item comes from the training generator (Section 2.2).

## 2 The model, the rows and the held-out items

Laya is a non-autoregressive decision model. It reads a message together with a typed question (a choice among named options, a level on an ordered scale, or a yes/no) and returns a probability for every option in a single forward pass, without writing any text. Its encoder is ModernBERT-large (28 layers, hidden size 1,024, about 395 million parameters). Each option is written into the input as a short description followed by one [MASK] marker.

The hidden state at each marker passes through a small two-layer head in which the option slots attend to each other, and a scorer turns every slot into one logit. A separate head estimates whether the model should act at all. The input is cut to 512 tokens, 192 of which are reserved for the question and its options.

The model is meant to be adapted. Its own trainer fine-tunes the whole network on rows of (message, question, expected answer) with a calibration-aware loss. It then re-fits per-type temperatures and abstention thresholds on a held-back slice and writes a new checkpoint. This paper treats that trainer as a fixed instrument and asks two questions about the *rows* that go into it. How many are needed before the held-out decisions stop improving? And how many of them may carry the wrong label before the decisions get worse?

### 2.1 Where the rows come from

A large language model (Qwen3.8-27B) wrote all 2,818 training rows and all held-out items, and the rows were then screened. For each (category, answer) pair the generator is asked for a batch of short, varied messages for which that answer is correct. A second, blind pass receives only the message and the option list and picks an option without knowing the intended answer; a row survives only if the blind answer equals the intended one. The surviving rows are shuffled and split into a training pool and held-out sets, and any held-out text that also occurs in the pool is removed. The pool covers nine categories: twelve-way support topics, hardware-failure triage, multilingual department routing, JSON-serialised tickets, a three-level urgency scale, and four small categories (injected instructions, negation, churn without an ultimatum, and long noisy messages).

Every label is a model opinion, so “noise” here means disagreement with the generator’s intent; no human labelled anything. The screening step also filters noise: a row whose blind answer differs from the intended one is dropped. It cannot remove an error that the second pass shares with the generator. Section 11 estimates how much label error is left in the held-out sets by comparing their gold labels with those of a second, independent model from another family.

**Table 1.** The training pool and the held-out items, grouped into the seven tasks used for per-task curves. “Option cells” counts (question schema, answer option) pairs; rows per cell is the mean over them and the minimum is the thinnest cell.

| Task | Training categories | Rows | Option cells | Rows / cell | Min cell | Held-out items |
|---|---|---|---|---|---|---|
| 12-way topics | many options | 1,145 | 12 | 95 | 30 | 270 |
| hardware failure | hw failure | 544 | 5 | 109 | 99 | 115 |
| urgency (3 levels) | urgency | 148 | 3 | 49 | 22 | 89 |
| JSON tickets | json state | 233 | 3 | 78 | 53 | 76 |
| multilingual dept. | multilingual | 399 | 3 | 133 | 128 | 346 |
| sentiment (injection) | injection | 109 | 3 | 36 | 30 | 127 |
| negation / churn / outage | negation, churn noul, long noisy | 240 | 7 | 34 | 29 | 42 |
| **all** | | **2,818** | | | | **1,065** |

**Table 2.** The seven held-out sets. None of their texts occurs in the training pool; they come from the same generator, so every accuracy below is an in-distribution accuracy. The last column is how often an independent labeller (another model family) gives the same answer as the gold label.

| Set | Items | Independent labeller agrees |
|---|---|---|
| probe (mixed) | 224 | 97.3% |
| sentiment (injection) | 106 | 99.1% |
| JSON tickets | 58 | 96.6% |
| hardware failure | 82 | 95.1% |
| 12-way topics | 195 | 94.4% |
| multilingual | 322 | 97.2% |
| urgency | 78 | 94.9% |

### 2.2 What training and test share

The training subsets and the held-out items come from one generator, one screening pass, the same nine categories and the same question schemas and answer options. They differ in the items: the texts were split once into a pool and held-out sets, and any held-out text that also occurs in the pool was removed (exact-text match only; near-duplicates were not searched for). Every subset is drawn from the pool alone, so no run sees a held-out item. Label noise is added to training labels only, and the held-out labels stay clean in every run, noisy ones included.

Two details need care. The calibration slice that the trainer uses for temperatures is cut from the training rows, never from the held-out items, so it holds no test data. Its rows count among the rows quoted here, though, and in the noisy runs it comes out of the corrupted file and carries the same share of wrong labels. The confusable-pair partner of each option was read from the clean models' probabilities on the held-out items and their gold labels. No trained model saw those items, but held-out information shaped the noise design. Section 3 says in which direction that can bias the comparison. Every accuracy in this paper is therefore an in-distribution accuracy on new items from the training generator. No test uses a different generator, other schemas, another phrasing style or real traffic. Section 13 treats that as the main limit on how far the curves transfer.

## 3 Design of the study

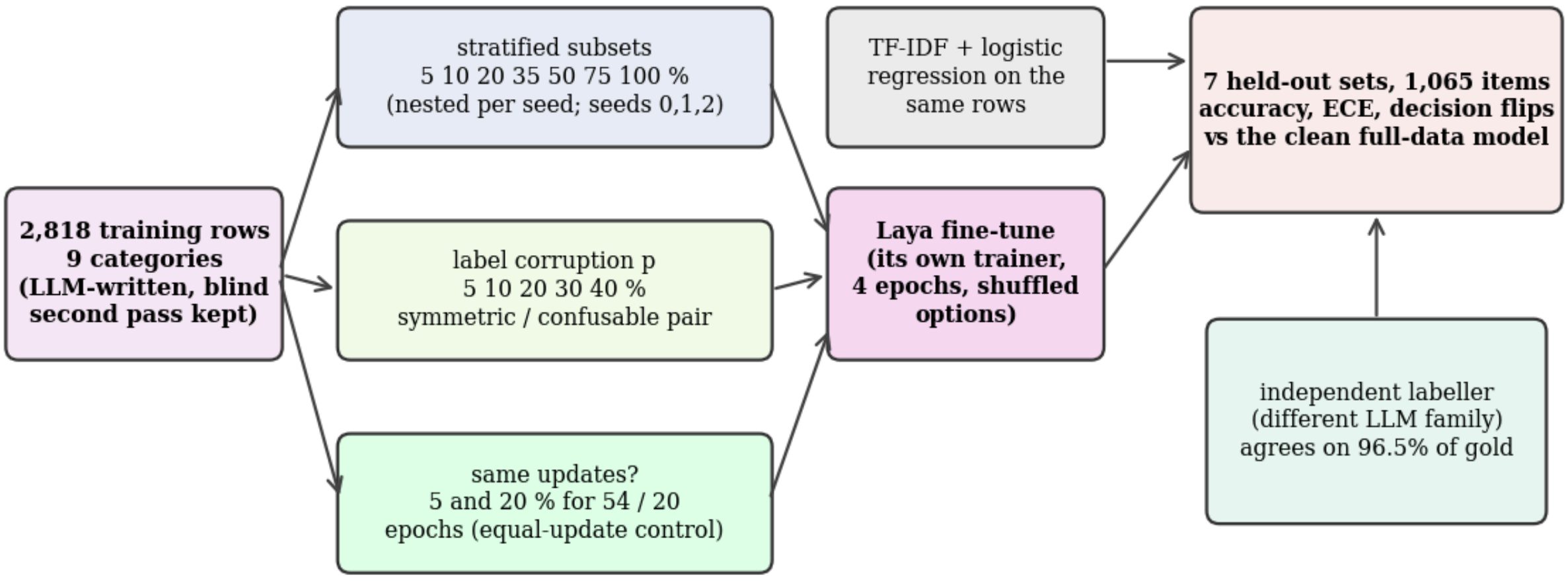

**Figure 1.** Study design. One pool of training rows is cut into nested stratified subsets, optionally corrupted at the label level, used to fine-tune the same encoder with the same trainer, and scored on seven held-out sets whose labels were never touched.

A subset of fraction $f$ keeps round($f \cdot m$) rows of every (category, answer) cell of size $m$, at least one, drawn by a seeded shuffle. Within one seed the subsets are nested (the 5% rows are inside the 10% rows and so on), so a curve isolates what the extra rows add. Fractions are 5, 10, 20, 35, 50 and 75%. Because the subset rounds per cell, they hold 143, 281, 564, 987, 1,413, 2,108 rows, a few off f·2,818, and the text quotes the actual counts. The 100% point is the full-data fine-tune, available for four seeds. The manifest in the code lists 35 possible fine-tunes, but because of a time budget only 15 were run: 8 clean subset runs, 2 equal-update controls and 5 label-noise runs, besides the four full-pool seeds (Appendix A lists each run). The 5% to 75% subsets have one seed except 20% and 50%, which have two; every noisy run has one seed.

In a run with noise level $p$ each retained row draws one uniform number from its own seeded generator, and its gold answer is replaced when the draw is below $p$, so the corrupted set at 5% is contained in the one at 10%. *Symmetric* noise replaces the answer by one of the other options chosen uniformly. *Confusable-pair* noise replaces it by that option's fixed partner, the wrong option that the four clean full-data models give the most probability mass on average (for an ordered scale, restricted to the adjacent levels, because those are the plausible mistakes of an annotator). This noise is class-conditional in the sense of the confident-learning literature and is the harder case, because the corrupted rows agree with each other. Two things differ between it and symmetric noise: the wrong labels of a class pile up on one option, and that option is a confusable one. No run isolates the second from the first, because a fixed random partner was not run. Only the training file is corrupted; no held-out label is changed, but the calibration slice is cut from the corrupted file, so the temperatures of a noisy run are fitted on labels that are partly wrong. The partner map was read off the held-out items, so it uses held-out *inputs and gold labels* of the clean models, but no held-out information ever enters a trained model. The map holds one choice per (question, option) pair, about thirty in all, so it carries little item-level information, and it was not re-derived from training data to measure the difference. If it matters, it matters in one direction: a map fitted on the test items can only make the confusable-pair noise more harmful to them, so the gap between confusable-pair and symmetric noise is, if anything, an overestimate.

Every run uses the recipe of the full-data model. The trainer's defaults apply (4 epochs, micro-batch 8 with 8-step gradient accumulation, encoder learning rate 2.5e-5, head learning rate 1e-4, the calibration-aware RLCD loss), with option order shuffled during training and a held-back calibration slice of 10% of the rows (at most 400). A run therefore trains on 129 of its 143 rows at 5% and on 2,537 of 2,818 at 100%. Only the rows and the seed change. Four epochs on fewer rows also means fewer optimizer updates (12 at 5% of the rows, 160 at 100%), so a size curve at fixed epochs confounds data with optimisation; the equal-update control in Section 5 separates the two. Runs share a GPU with other jobs and weights were kept in memory, so wall-clock times are not reported.

The metrics are accuracy on the pooled 1,065 held-out items (and per task), expected calibration error with 15 equal-width bins, mean confidence and decision flips (the share of held-out items whose top answer differs from the clean full-data seed-0 model). Intervals on a single run are 95% Wilson intervals; paired comparisons on identical items use the exact McNemar test.

Accuracy against fraction is fitted with $\mathrm{err}(f) = e_\infty + a \cdot f^{-b}$ by least squares over a grid of $b$ in [0.05, 3], with $e_\infty \geq 0$ and $a \geq 0$, on the seed means at the seven fractions. Uncertainty comes from a 300-draw bootstrap that resamples held-out items and (within each fraction) seeds before refitting. The plateau, the rows needed to reach 95% and 99% of the plateau, and the marginal gain of doubling are read off every refit. A saturating form replaces a pure power law because a classifier on a fixed label set cannot pass the ceiling of its test labels. With seven points and three parameters the fit summarises the curve; its intervals show how loosely it is pinned.

## 4 How accuracy grows with rows

Without any fine-tuning the shipped model answers 69.4% of the 1,065 held-out items correctly (Figure 2, Table 3). The smallest subset, 5% of the pool or 143 rows (about 4 per question-option cell), already lifts that to 74.3%, and the full pool of 2,818 rows reaches 86.9% (s.d. 0.1 over four seeds). The first 143 rows therefore buy 28% of the total improvement and the remaining 2,675 rows buy the rest. Seeds per size are 5%: 1, 10%: 1, 20%: 2, 35%: 1, 50%: 2, 75%: 1, 100%: 4.

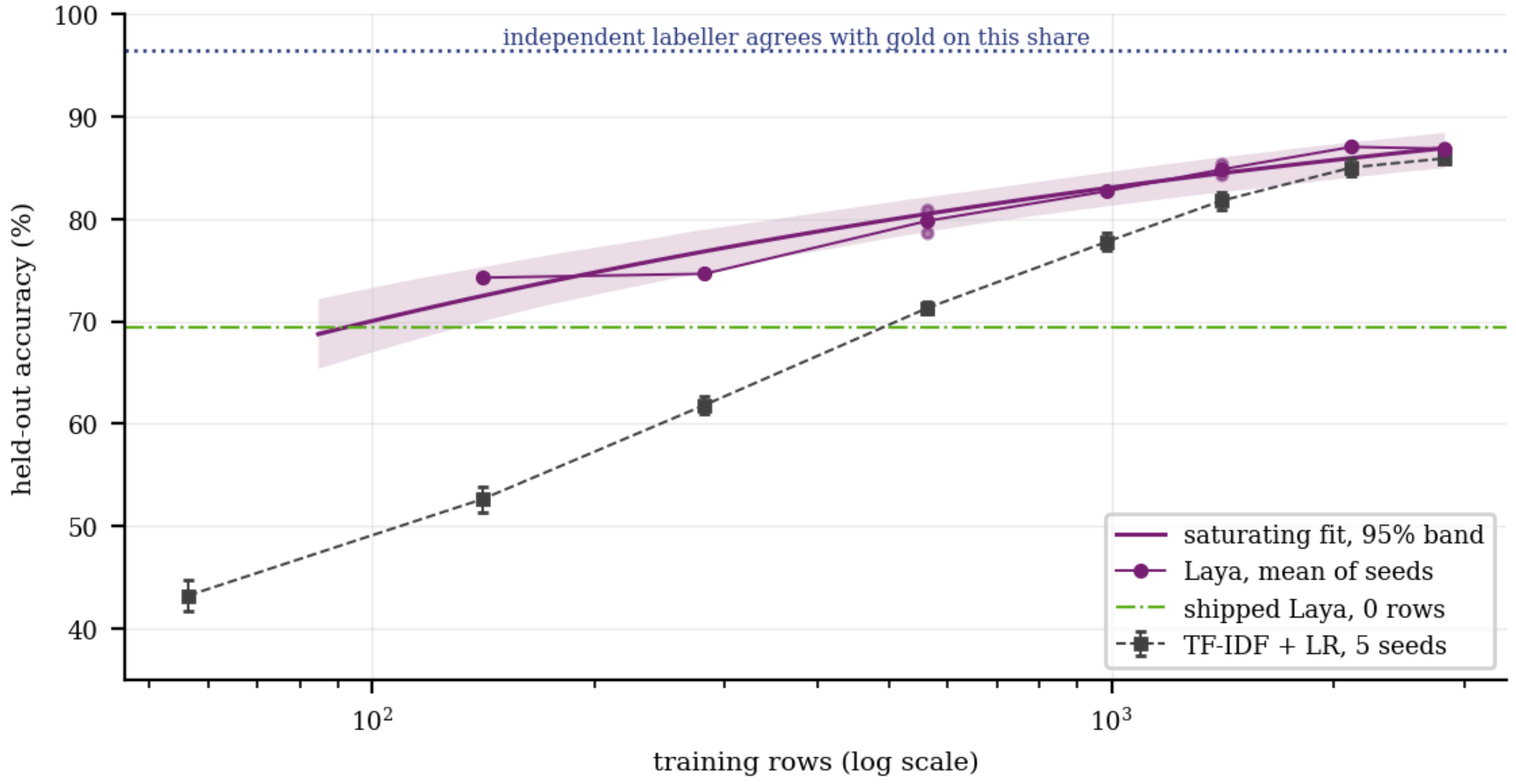

**Figure 2.** Pooled held-out accuracy (n=1,065 items) against training rows. Dots are single fine-tuning runs, the line is the mean of seeds, the band is the 95% bootstrap interval of the fitted error = $e_\infty + a \cdot f^{-b}$. The dotted line is how often a second, independent LLM labeller agrees with the gold labels.

**Table 3.** Pooled held-out accuracy (%) against the share of training rows. "runs" is the number of seeds; "step" is the change in the seed mean from the previous row (points); TF-IDF is a logistic regression on word and character n-grams, mean ± s.d. of 5 seeds (the 100% row is deterministic); the last column is Laya minus TF-IDF trained on the same rows, with a 95% bootstrap interval over the 1,065 items.

| Rows used | rows | runs | updates | Laya mean ± s.d. | range | step | ECE | TF-IDF + LR | Laya – TF-IDF, same rows |
|---|---|---|---|---|---|---|---|---|---|
| 5% | 143 | 1 | 12 | 74.3 | n/a | | 6.7 | 52.6 ± 1.2 | +21.1 [+17.6, +24.5] |
| 10% | 281 | 1 | 16 | 74.6 | n/a | +0.4 | 9.6 | 61.8 ± 0.8 | +12.5 [+9.2, +16.2] |
| 20% | 564 | 2 | 32 | 79.8 ± 1.6 | 78.7-80.9 | +5.2 | 4.1 | 71.3 ± 0.6 | +8.9 [+6.4, +11.5] |
| 35% | 987 | 1 | 56 | 82.7 | n/a | +2.9 | 5.2 | 77.8 ± 0.8 | +5.1 [+2.2, +8.0] |
| 50% | 1,413 | 2 | 80 | 84.8 ± 0.7 | 84.3-85.4 | +2.1 | 6.2 | 81.8 ± 0.9 | +3.8 [+1.2, +6.2] |
| 75% | 2,108 | 1 | 120 | 87.0 | n/a | +2.2 | 6.1 | 85.0 ± 0.9 | +1.3 [-1.2, +3.7] |
| 100% | 2,818 | 4 | 160 | 86.9 ± 0.1 | 86.8-86.9 | -0.2 | 7.0 | 85.9 ± 0.0 | +0.9 [-1.5, +3.4] |

Most of the curve is a rise followed by a shoulder. From 35% (987 rows) to 100% the seed mean moves by +4.1 points, and from 75% to 100% by -0.2 points. The binomial standard error of one run on 1,065 items is 1.0 points, so the last step is inside twice that error. The 75% point is a single seed. The saturating fit has a plateau of 99.9% with a 95% bootstrap interval [97.8-100.0]. A fitted plateau at the 100% ceiling means that the fit found no error floor. That cannot be right, because the independent labeller bounds the attainable accuracy near 97%, so the fitted tail describes the slope near the pool size and is no forecast. The data cannot tell plateaus apart. With the plateau pinned at 94%, 96% or 98% the fit leaves a root-mean-square residual of 1.34, 1.28 and 1.28 points on the seven seed means, against 1.19 for the free fit, while the binomial error of one run is 1.0 points. The bootstrap interval covers sampling of items and seeds inside one functional form and understates this. The reliable statement is local. On the fit, 100 more rows on top of the full pool are worth +0.11 points (interval +0.09 to +0.12), and gaining two more points would take 2.0× [1.9-2.2] the current pool (Table 7). Figure 3 turns the same fit into the gain from doubling: +2.5 points when going from 1,200 to 2,400 rows, with a bootstrap interval from +2.1 to +2.8. Doubling again beyond the pool is an extrapolation and is drawn in grey.

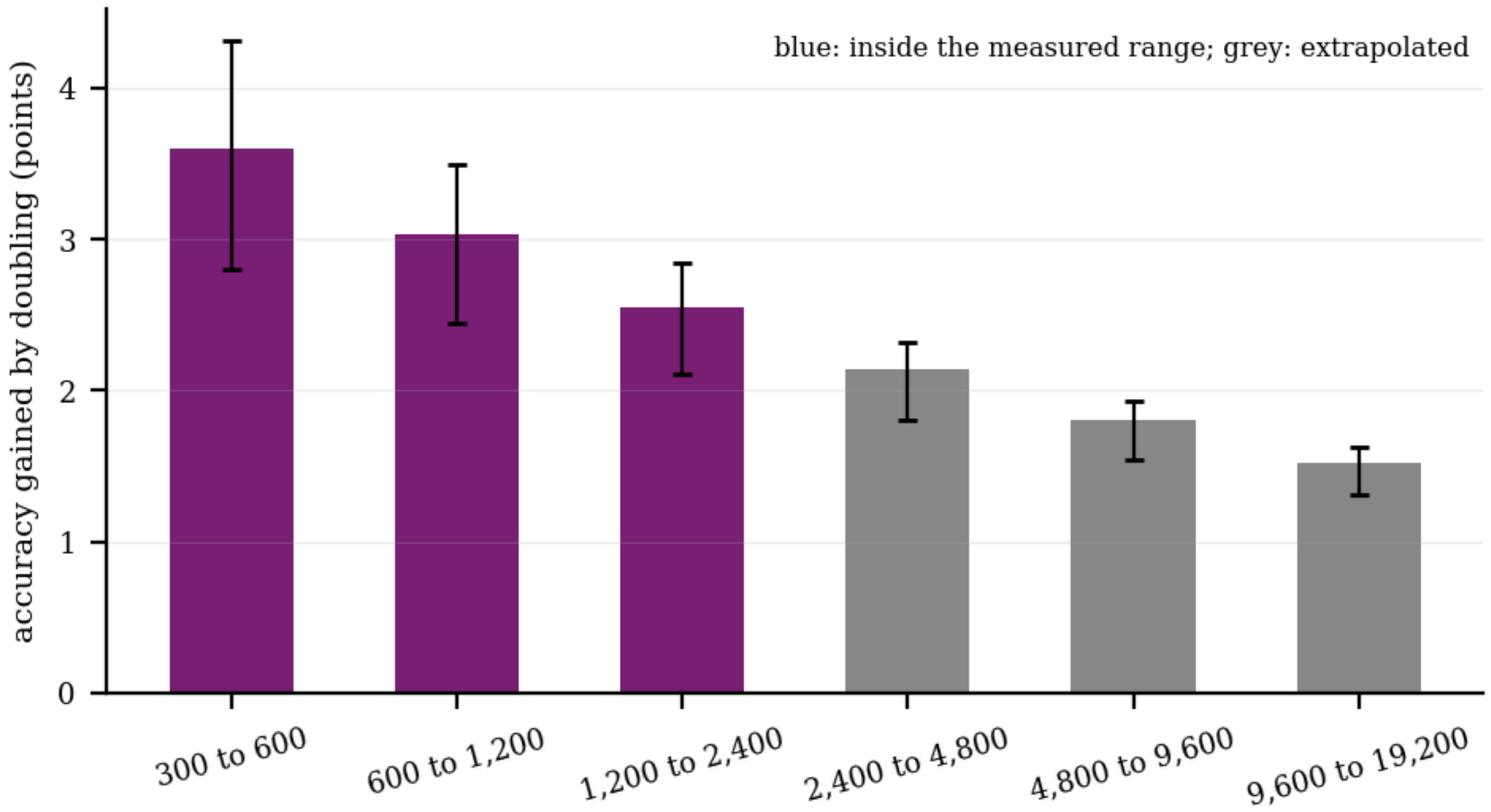

**Figure 3.** What doubling the number of rows is predicted to buy on the pooled curve, with bootstrap intervals of the fit. Bars beyond 2,818 rows (grey) are extrapolation and should be read as a hypothesis.

### 4.1 Seed noise against the steps between sizes

Fine-tuning on the same rows with a different seed changes the pooled accuracy by only a few tenths of a point at 100% (the four full-pool seeds span 86.8-86.9%). The spread is larger when the data are small. Table 4 lists the seed standard deviation at each size beside the binomial standard error of a single run and the step to the next size, and Figure 4 plots the same comparison. A step smaller than twice the seed standard deviation is marked “within seed noise”, and no slope should be read off it. Where only one seed exists (every size except 20%, 50% and 100%) no standard deviation can be computed, so Table 4 lists only the steps that start at a size with at least two seeds.

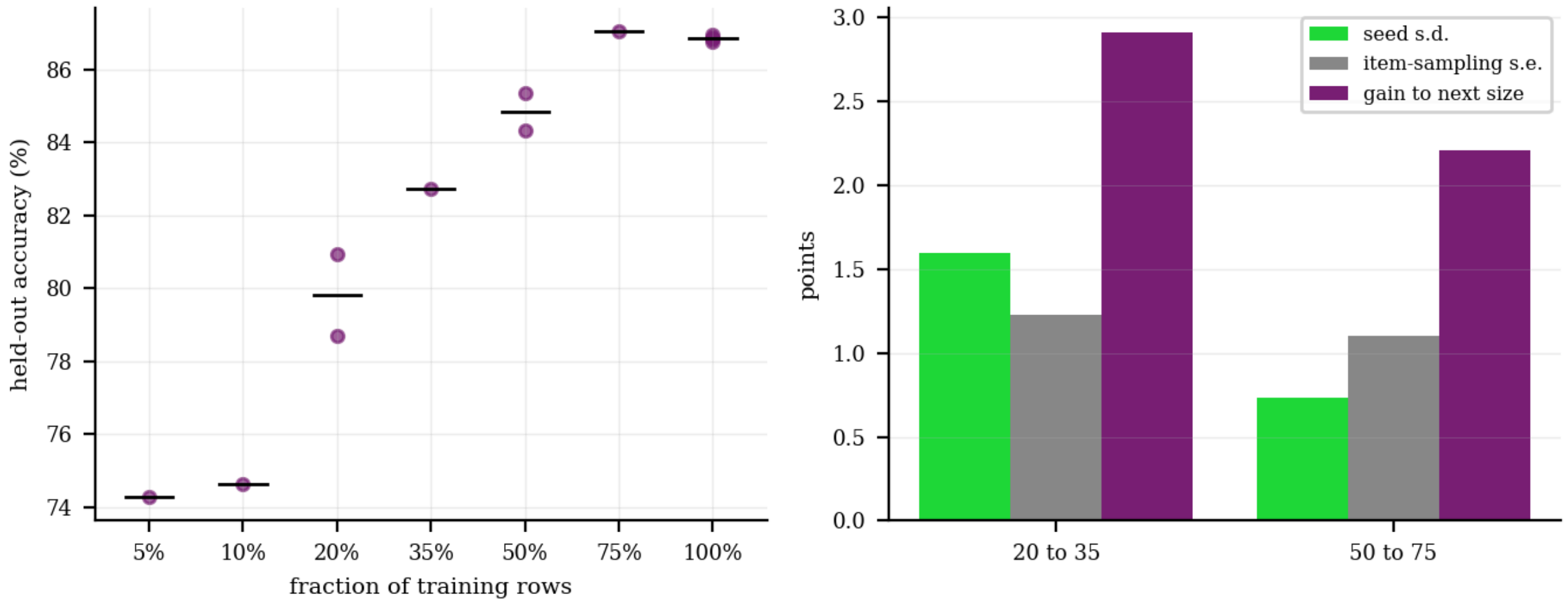

**Figure 4.** Left: single runs (dots) and their mean. Right: spread between seeds and the binomial standard error of one run, set against the gain that the next data step buys. Where the orange bar is as tall as the blue one the step is within seed noise.

**Table 4.** Is a data step bigger than the noise between seeds? s.d. is across seeds at the smaller size, s.e. the binomial standard error of one run on the 1,065 items; the verdict compares the step in the seed mean with twice the s.d.

| Step | seeds | seed s.d. | binomial s.e. | step | verdict |
|---|---|---|---|---|---|
| 20% to 35% | 2 | 1.59 | 1.23 | +2.9 | within seed noise |
| 50% to 75% | 2 | 0.73 | 1.10 | +2.2 | step larger than 2 s.d. |

## 5 Rows or updates? An equal-update control

With four epochs fixed, the 5% subset receives 12 optimizer updates and the full pool 160, so a learning curve at fixed epochs mixes two causes: fewer distinct rows, and less optimisation. To separate them, the 5% and 20% subsets were re-trained for 54 and 20 epochs, which gives each of them about 160 updates (the update count is not exactly 160 because each epoch is rounded up to whole accumulation steps). Every other setting is unchanged.

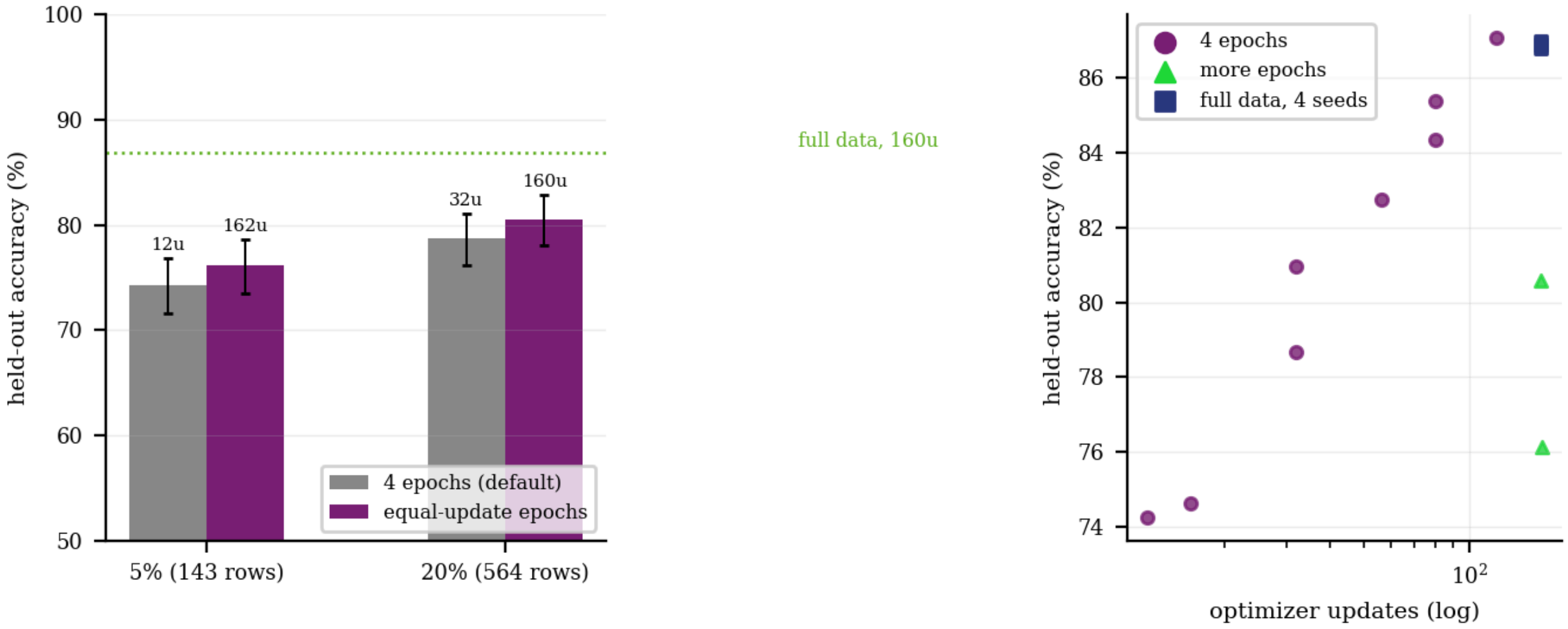

**Figure 5.** Left: with the number of optimizer updates raised to the full-data level (labels above the bars: updates), the small subsets are re-trained for more epochs; error bars are 95% Wilson intervals. Right: every clean run, accuracy against updates (effective batch 64).

**Table 5.** Equal-update control (seed 0). The same rows are trained for as many epochs as it takes to reach about 160 optimizer updates, the number the full pool receives at 4 epochs. “gap closed” is the share of the distance to the mean full-data accuracy that the extra epochs recover.

| Rows | updates 4ep / long | 4 epochs: accuracy [95% CI] | epochs: accuracy [95% CI] | difference (McNemar) | ECE 4ep / long | gap closed |
|---|---|---|---|---|---|---|
| 5% (143 rows) | 12 / 162 | 74.3% [71.6-76.8] | 54: 76.2% [73.5-78.6] | +1.9 (p=0.145) | 6.7 / 20.6 | 15% |
| 20% (564 rows) | 32 / 160 | 78.7% [76.1-81.0] | 20: 80.6% [78.1-82.8] | +1.9 (p=0.130) | 5.3 / 13.1 | 23% |

At 5% of the rows (143), training 54 epochs (default: 4) moves accuracy from 74.3% to 76.2% (+1.9 points, McNemar p=0.145), which closes 15% of the distance to the full-pool mean of 86.9%. At 20% of the rows (564), training 20 epochs (default: 4) moves accuracy from 78.7% to 80.6% (+1.9 points, McNemar p=0.130), which closes 23% of the distance to the full-pool mean of 86.9%. The second four-epoch seed at 20% of the rows reaches 80.9%, 0.4 points above the 20-epoch run, so the gain at 20% is smaller than the 2.3-point spread between two ordinary seeds and cannot be told from seed noise. Most of the gap stays open, which fits the rows being the limit (there are fewer of them, and they cover each option less often), but this control cannot show it. Extra epochs on 143 or 564 rows repeat the same examples. The control is one seed per size and one recipe, and neither gain is significant on its own (McNemar p above 0.10 in both rows of Table 5), so it bounds the optimisation effect without measuring it. Longer training closes at most a quarter of the gap to the full pool and comes with a large rise in ECE. Whether it is worth it for a small labelled set has to be judged on held-out accuracy and calibration, and the training loss cannot settle it. Longer training on few rows changes ECE (Table 5) from 6.7 to 20.6 at 5% and from 5.3 to 13.1 at 20%.

## 6 Calibration at small sizes

Fine-tuning data also set how well the stated probabilities can be trusted. Figure 6 shows the expected calibration error across sizes and Figure 7 reliability diagrams at three of them. The trainer re-fits temperatures on a held-back slice of the rows, so the slice shrinks with the subset: at 5% it is 14 items, which is small for fitting per-type temperatures.

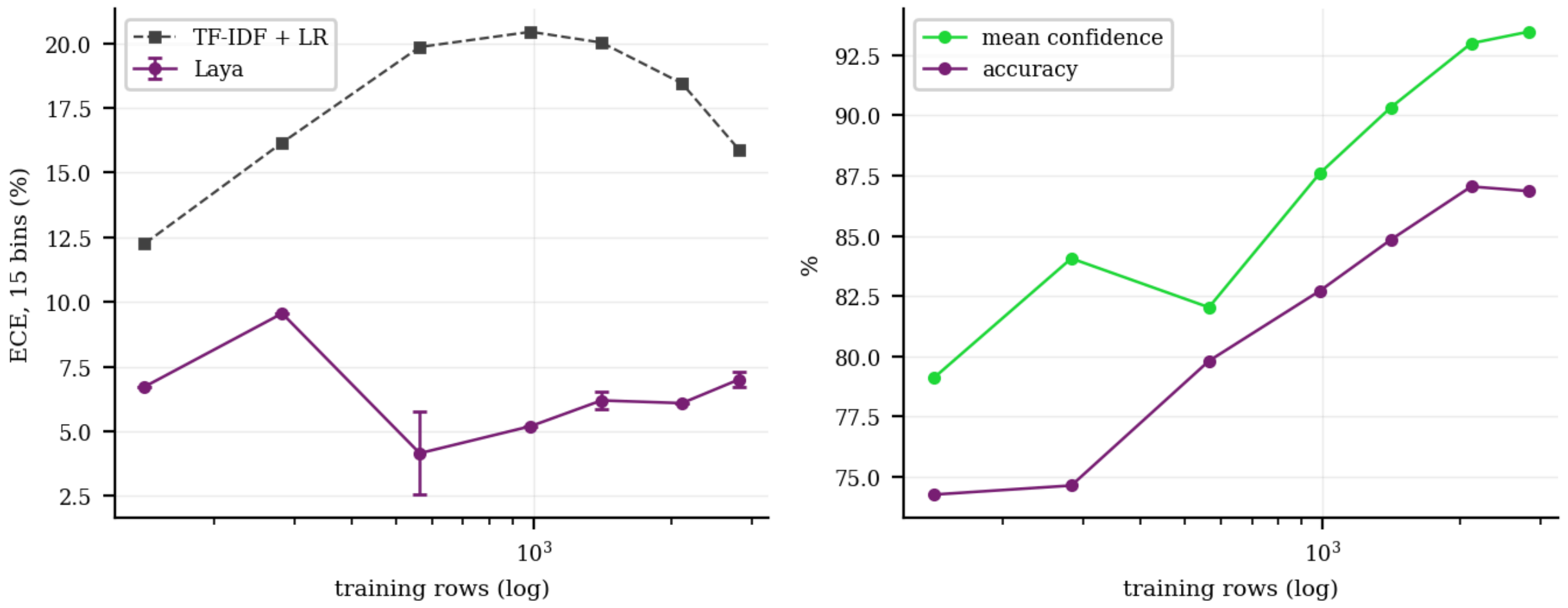

**Figure 6.** Calibration against data size. Left: expected calibration error (mean and s.d. over seeds). Right: mean top-probability and accuracy; the gap between them is the over-confidence that the left panel summarises.

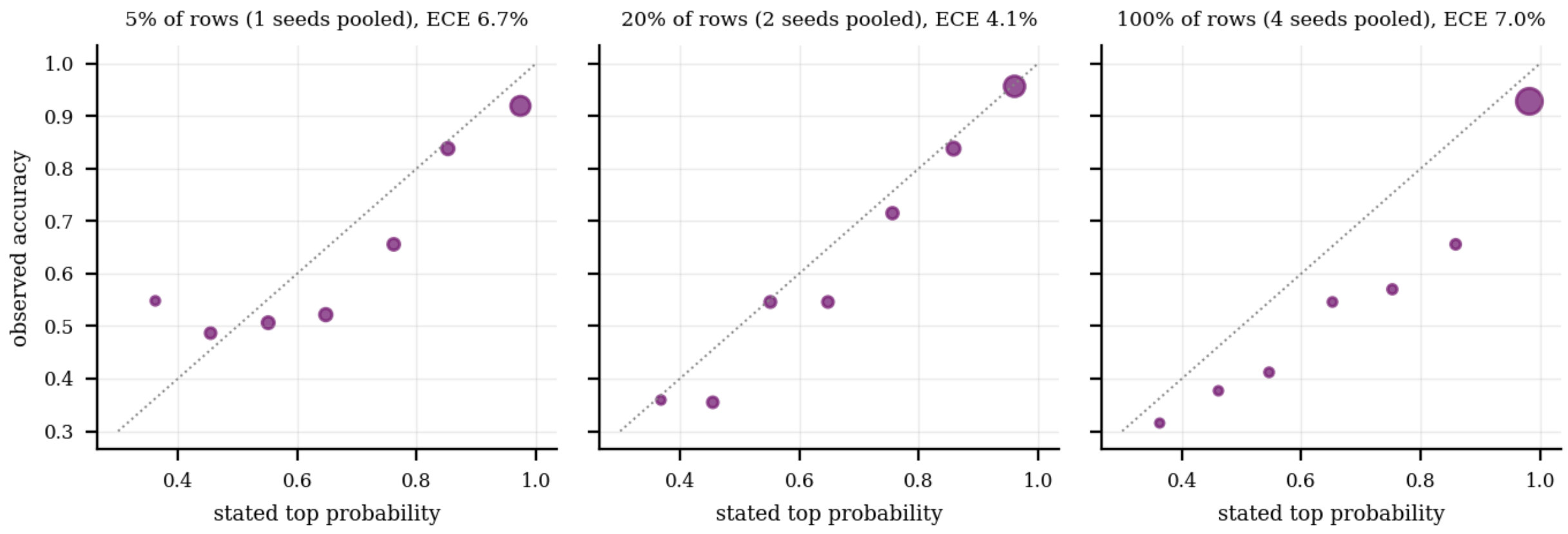

**Figure 7.** Reliability of the stated probability at three data sizes (runs pooled over seeds; marker area is the share of items in the bin; bins with fewer than 9 items omitted).

The ECE is highest at 10% of the rows (9.6%) and lowest at 20% (4.1%); it does not fall monotonically with data. Mean confidence minus accuracy is +4.8 points at 5% and +6.6 points at 100%, so the model is over-confident throughout. The bag-of-n-grams classifier has an ECE of 12.3% at 5% and 15.9% at 100%. Calibration differences between sizes are of the same order as the seed spread (Figure 6, error bars), so a calibration check should use a held-out set from the deployment distribution, because the size of the training set does not settle it.

## 7 Which tasks need the data

Pooled accuracy hides tasks that differ a great deal. Figure 8 plots the same learning curve for each of the seven tasks against the number of rows of that task, and Table 6 gives the seed means. The shipped model (red dash-dot lines) is already at 92% on the sentiment task with injected instructions and at 93% on the three small categories, but at 58% on urgency, 66% on hardware and 60% on multilingual routing.

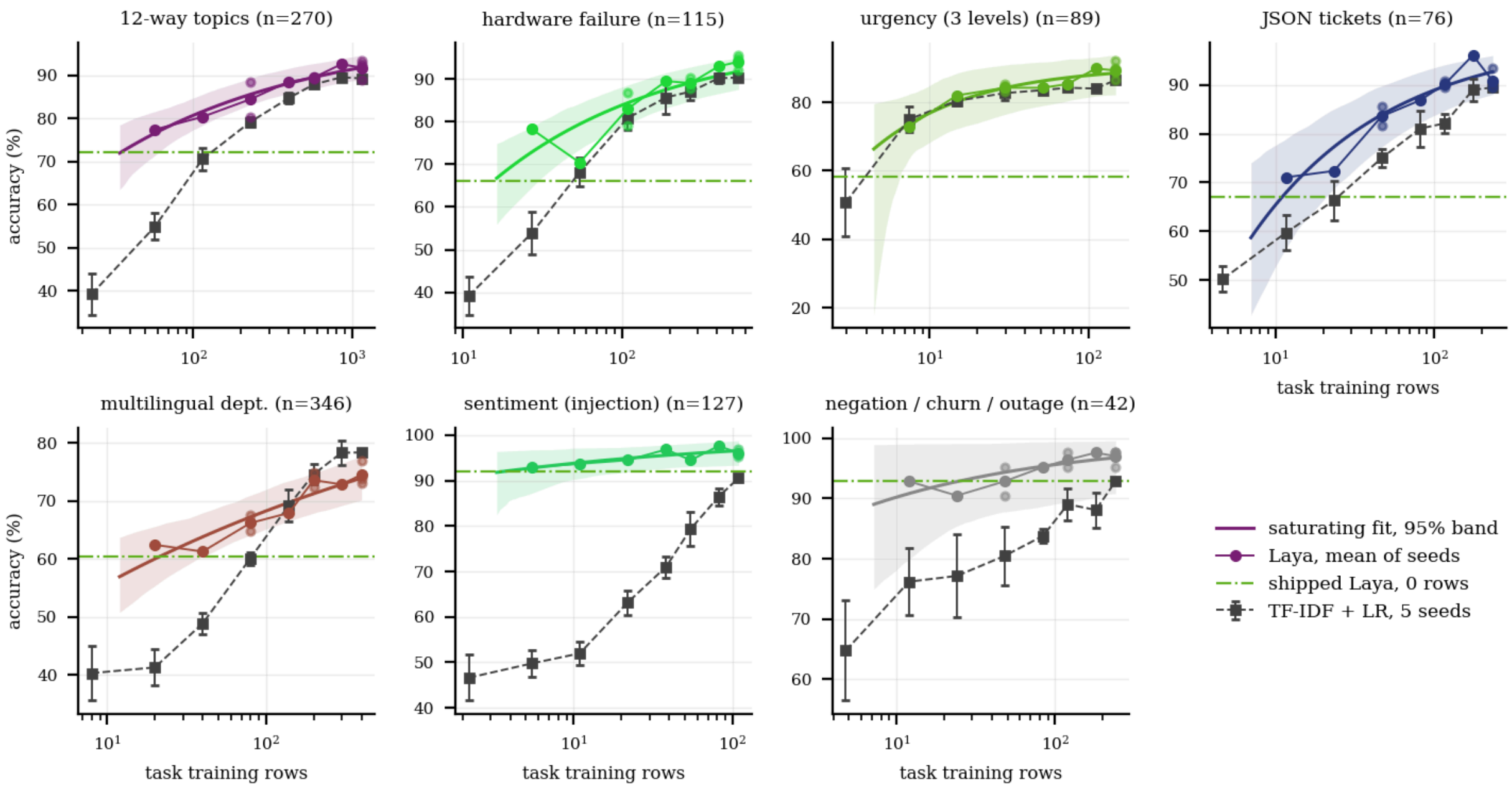

**Figure 8.** The same curve for each task, on that task's own training rows (x-axis) and held-out items (title). Bands are 95% bootstrap intervals over held-out items and seeds; the dashed grey line is TF-IDF + logistic regression on the same fractions.

**Table 6.** Task accuracy (%, seed means) at each fraction; "s.d." is the spread over the four full-data seeds; the last column is TF-IDF + LR on the full pool.

| Task | 5% | 10% | 20% | 35% | 50% | 75% | 100% | s.d. @100% | TF-IDF @100% |
|---|---|---|---|---|---|---|---|---|---|
| 12-way topics | 77 | 80 | 84 | 89 | 89 | 93 | 92 | 2.0 | 89 |
| hardware failure | 78 | 70 | 83 | 90 | 89 | 93 | 94 | 1.5 | 90 |
| urgency (3 levels) | 73 | 82 | 84 | 84 | 85 | 90 | 89 | 2.3 | 87 |
| JSON tickets | 71 | 72 | 84 | 87 | 90 | 96 | 91 | 1.9 | 89 |
| multilingual dept. | 62 | 61 | 66 | 68 | 74 | 73 | 74 | 1.7 | 78 |
| sentiment (injection) | 93 | 94 | 94 | 97 | 94 | 98 | 96 | 0.6 | 91 |
| negation / churn / outage | 93 | 90 | 93 | 95 | 96 | 98 | 97 | 1.2 | 93 |

Figure 9 divides each cell by the task's full-pool accuracy. With 5% of the rows the task that keeps the least is JSON tickets (78% of its full accuracy), the task that keeps the most is sentiment (injection) (97%). At 20% of the rows hardware failure, multilingual dept. still sit below 90% of their final accuracy. The multilingual task is a special case: at 100% of the rows the TF-IDF classifier (78.3%) is ahead of the fine-tuned encoder (74.5%). That is consistent with a backbone whose pre-training is English and whose subword vocabulary splits other scripts into many pieces. More rows of the same kind help the n-gram model more than they help Laya: from 5% to 100% of the rows it gains 37 points on this task against 12 for the encoder.

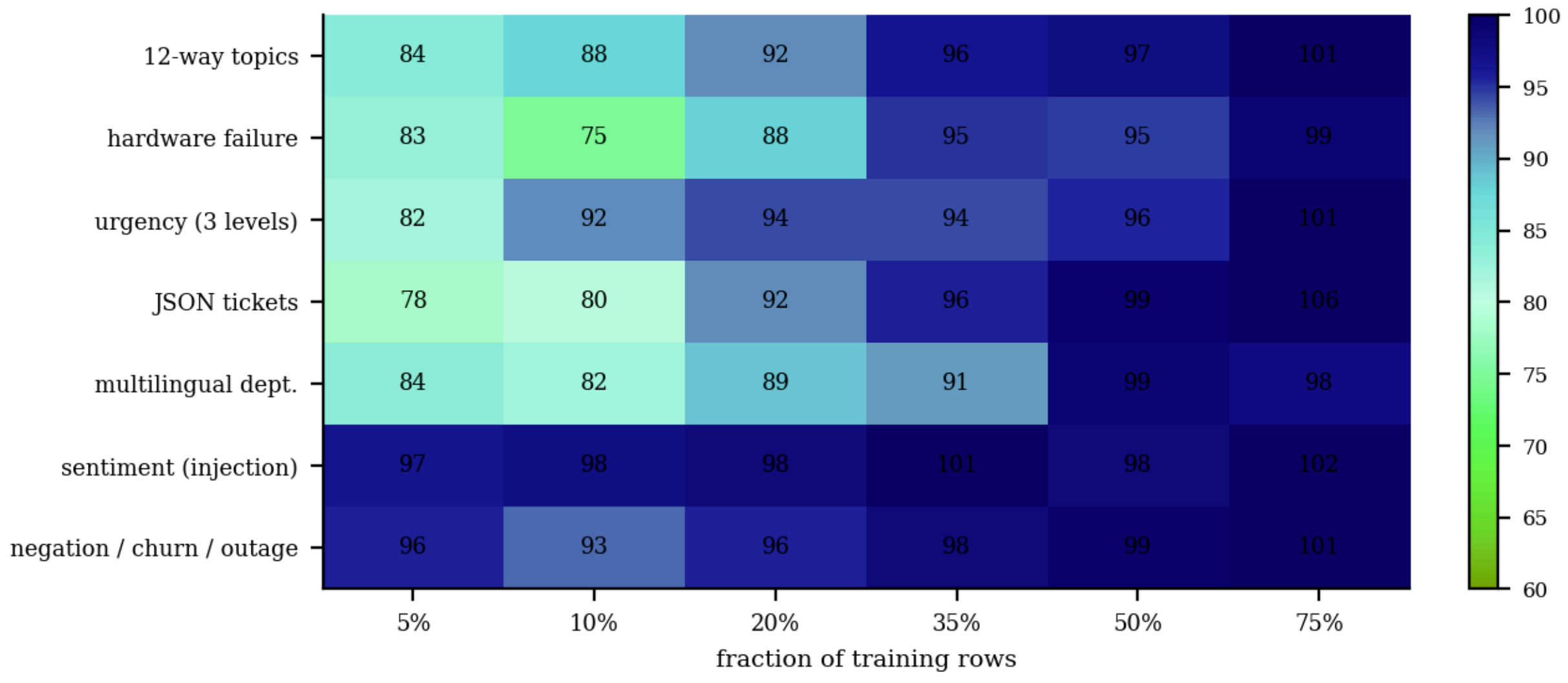

**Figure 9.** Share of full-data accuracy retained (%, seed means) by each task at each data fraction. Rows that stay green from 20% onwards are the tasks that do not need the whole pool; the red cells are where the next rows pay off most.

Is the demand for rows a property of the task or only of how many rows it already has? Figure 10 plots retention at 20% of the rows against the rows per answer option in the full pool. The rank correlation is -0.93 over seven tasks. Its sign is the opposite of “more rows per option, more retained”. The two tasks that stay below 90% at 20% of the rows (hardware failure, multilingual dept.) have the second and first thickest pools (109 and 133 rows per cell), and the thinnest pools (negation / churn / outage, sentiment) are the easiest tasks. Pool thickness and difficulty are confounded here, and seven points cannot separate them, so the correlation says nothing about whether more rows per option would help a given task.

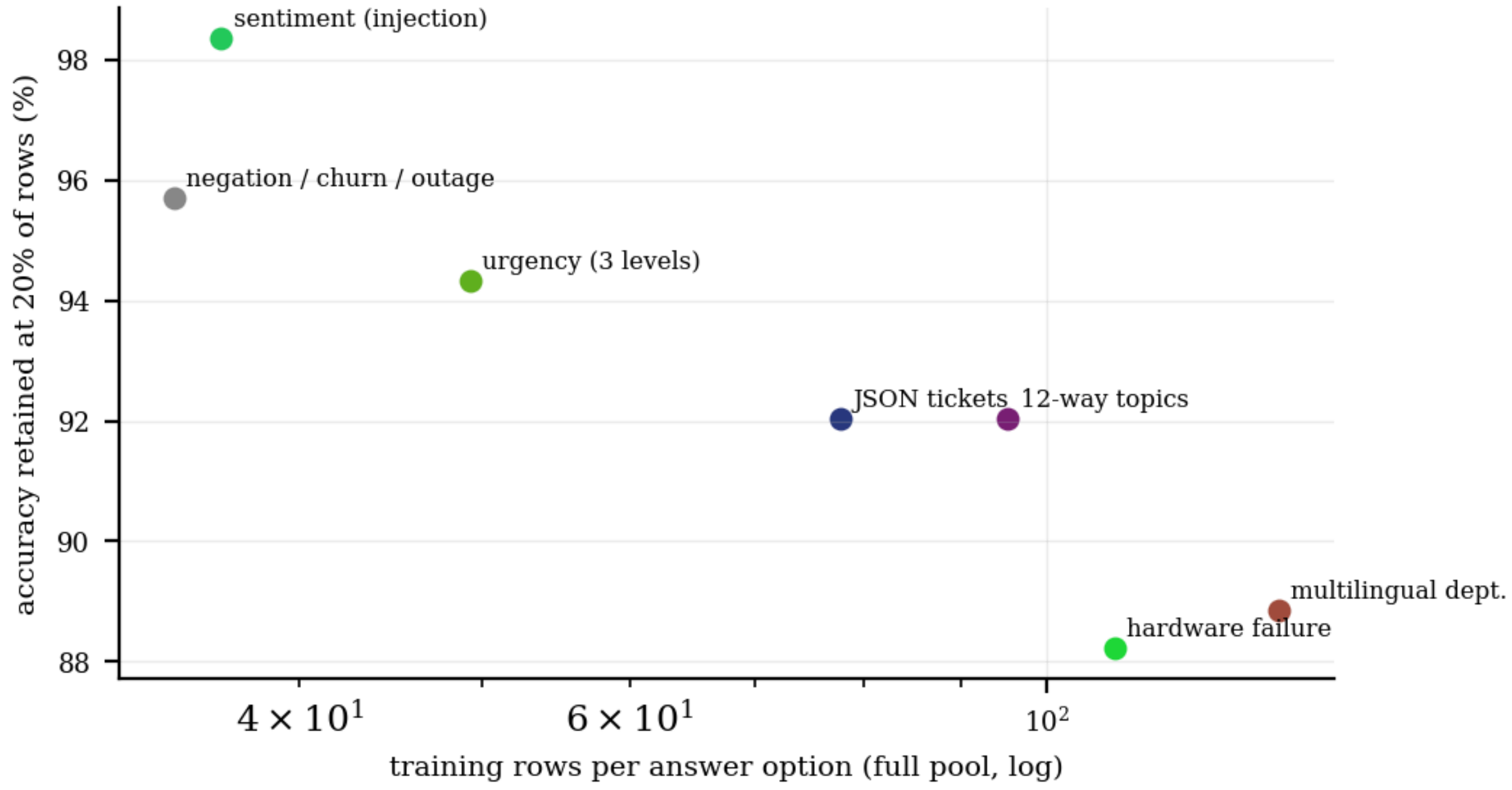

**Figure 10.** Retention at 20% of the rows against the number of training rows each answer option has in the full pool (Spearman ρ = -0.93, n=7 tasks, so only a trend).

## 7.1 Rows needed for further gains

The fit gives a hypothesis for each task about how much the pool would have to grow for further gains (Figure 11, Table 7). Because the plateaus are not pinned down, the useful quantity is the multiple of the current pool needed to add one or two points. The bootstrap intervals are wide, often unbounded. They are narrowest for the tasks with many held-out items and many rows. Treat the right-hand tail as unknown.

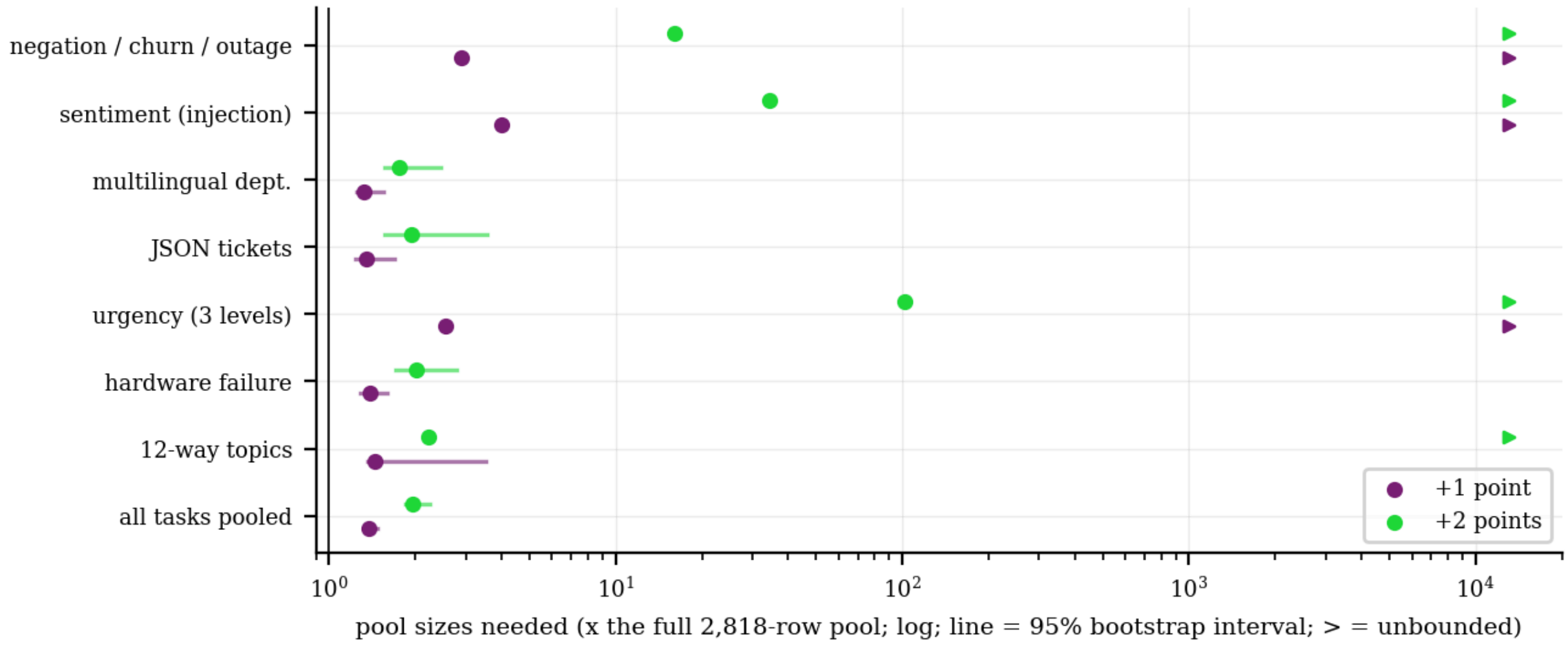

**Figure 11.** How many times the current pool the fitted curve says is needed to gain one (blue) or two (orange) accuracy points over the full-pool fit, per task, with 95% bootstrap intervals. Intervals that run off the right edge cannot rule out that the curve has already flattened.

**Table 7.** Saturating fit err = $e_\infty + a \cdot f^{-b}$ to the seed means. Plateau is $100 \cdot (1 - e_\infty)$ with its 95% bootstrap interval. The next two columns say by how many times the pool would have to grow for the fitted curve to gain one or two more points ("unbounded": beyond 10,000× or beyond the fitted plateau); the last column is the fitted gain in points from 100 extra rows on top of the full pool. Intervals are 2.5-97.5% of 300 bootstrap refits.

| Scope | pool rows | plateau (%) | b | pool size for +1 point | pool size for +2 points | points per +100 rows |
|---|---|---|---|---|---|---|
| **pooled** | 2818 | 99.9 [97.8-100.0] | 0.25 | 1.4× [1.3-1.5] | 2.0× [1.9-2.2] | +0.11 [+0.09, +0.12] |
| 12-way topics | 1145 | 100.0 [92.4-100.0] | 0.35 | 1.4× [1.4-3.5] | 2.2× [2.0-unbounded] | +0.24 [+0.10, +0.28] |
| hardware failure | 544 | 100.0 [99.8-100.0] | 0.40 | 1.4× [1.3-1.6] | 2.0× [1.7-2.8] | +0.53 [+0.39, +0.68] |
| urgency (3 levels) | 148 | 90.6 [83.3-100.0] | 0.70 | 2.5× [1.3-unbounded] | 102× [1.8-unbounded] | +0.63 [+0.00, +1.77] |
| JSON tickets | 233 | 99.8 [98.6-100.0] | 0.50 | 1.4× [1.2-1.7] | 1.9× [1.6-3.6] | +1.16 [+0.71, +1.64] |
| multilingual dept. | 399 | 98.4 [91.3-100.0] | 0.15 | 1.3× [1.2-1.6] | 1.8× [1.6-2.5] | +0.81 [+0.51, +1.02] |
| sentiment (injection) | 109 | 100.0 [94.3-100.0] | 0.25 | 4.0× [2.2-unbounded] | 34× [6.4-unbounded] | +0.51 [+0.00, +0.83] |
| negation / churn / outage | 240 | 100.0 [95.0-100.0] | 0.35 | 2.9× [1.7-unbounded] | 16× [3.0-unbounded] | +0.37 [+0.00, +0.71] |

**Table 8.** Practical budget per task, in training rows of that task (rows per question-option cell in brackets). "Measured" columns are the smallest tested fraction whose seed-mean accuracy is within 5 or 2 points of the full-pool accuracy; they need no curve fit but are limited to the tested grid and to 42-346 held-out items per task. The fitted columns extrapolate: the multiple of the pool needed for two more points, and the gain from 100 extra rows, with 95% bootstrap intervals.

| Task | rows in the pool | accuracy at 100% (%) | measured: within 5 points | measured: within 2 points | fitted: pool size for +2 points | fitted: points per +100 rows |
|---|---|---|---|---|---|---|
| 12-way topics | 1,145 (95/cell) | 91.8 | 401 (33/cell) | 859 (72/cell) | 2.2× [2.0-unbounded] | +0.24 [+0.10, +0.28] |
| hardware failure | 544 (109/cell) | 94.1 | 190 (38/cell) | 408 (82/cell) | 2.0× [1.7-2.8] | +0.53 [+0.39, +0.68] |

| Task | rows in the pool | accuracy at 100% (%) | measured: within 5 points | measured: within 2 points | fitted: pool size for +2 points | fitted: points per +100 rows |
|---|---|---|---|---|---|---|
| urgency (3 levels) | 148 (49/cell) | 89.3 | 74 (25/cell) | 111 (37/cell) | 102× [1.8-unbounded] | +0.63 [+0.00, +1.77] |
| JSON tickets | 233 (78/cell) | 90.8 | 82 (27/cell) | 116 (39/cell) | 1.9× [1.6-3.6] | +1.16 [+0.71, +1.64] |
| multilingual dept. | 399 (133/cell) | 74.5 | 200 (66/cell) | 200 (66/cell) | 1.8× [1.6-2.5] | +0.81 [+0.51, +1.02] |
| sentiment (injection) | 109 (36/cell) | 96.1 | 5 (2/cell) | 22 (7/cell) | 34× [6.4-unbounded] | +0.51 [+0.00, +0.83] |
| negation / churn / outage | 240 (34/cell) | 97.0 | 12 (2/cell) | 84 (12/cell) | 16× [3.0-unbounded] | +0.37 [+0.00, +0.71] |

## 8 Against a bag-of-n-grams classifier

To judge whether the pre-trained encoder is worth its price, we compare it with a cheap alternative on the same rows. A TF-IDF classifier (word and character n-grams, one logistic regression per question schema, C = 10) was trained on exactly the subsets and the corrupted labels used for Laya. Its probability is the softmax output of the regression, so its ECE can be compared too.

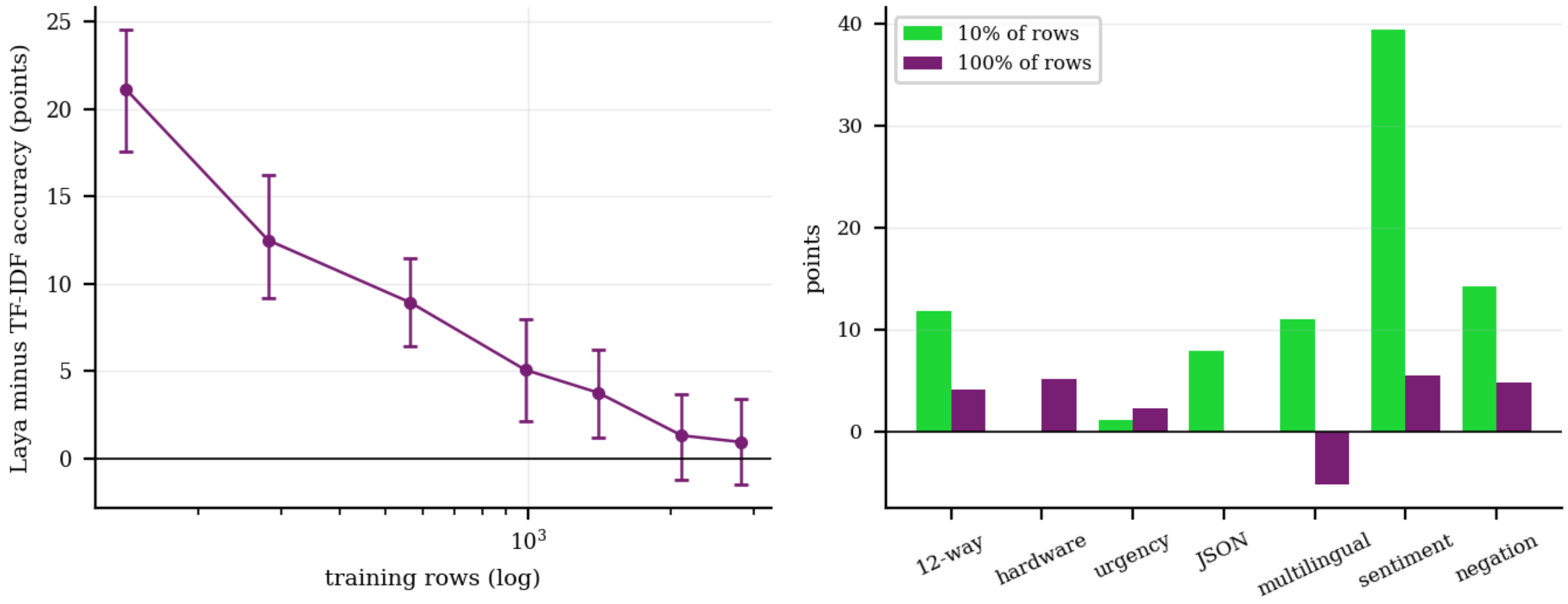

**Figure 12.** Laya minus a TF-IDF + logistic-regression classifier trained on exactly the same rows. Left: pooled gap with a 95% bootstrap interval over held-out items (seed-averaged where seeds exist). Right: the gap per task at 10% and 100% of the rows (seed 0).

At 5% of the rows Laya is ahead by +21.1 points (bootstrap interval +17.6 to +24.5); at 100% the difference is +0.9 points (-1.5 to +3.4). The interval of the gap includes zero from 75% of the rows upward. Table 9 reads the same result as a sample-efficiency ratio: the number of rows the n-gram model needs to reach the accuracy that Laya has with a given subset. The advantage is therefore a small-data advantage. Laya reaches a given accuracy with far fewer rows, and once a few thousand rows of this kind exist it is no more accurate than the n-gram model. For the multilingual task the n-gram model overtakes the encoder (right panel of Figure 12 shows the per-task gap at 10% and 100%).

**Table 9.** Sample efficiency against a bag-of-n-grams classifier: how many rows the TF-IDF + LR mean curve needs (log-linear interpolation between its measured sizes) to reach the accuracy that Laya has with the rows in the first column.

| Laya rows | rows | Laya accuracy (%) | TF-IDF rows to match | ratio |
|---|---|---|---|---|
| 5% | 143 | 74.3 | 728 | 5.1× |
| 10% | 281 | 74.6 | 752 | 2.7× |
| 20% | 564 | 79.8 | 1,182 | 2.1× |

| Laya rows | rows | Laya accuracy (%) | TF-IDF rows to match | ratio |
|---|---|---|---|---|
| 35% | 987 | 82.7 | 1,584 | 1.6× |
| 50% | 1,413 | 84.8 | 2,062 | 1.5× |
| 75% | 2,108 | 87.0 | not reached by 2,818 | > 1.3× |

## 9 How wrong can the labels be?

Figure 13 shows what the confusable-pair process, defined in Section 3, does on the twelve-way topic task. Each gold option is sent to the single option on which the clean models already place the most wrong-answer mass, so a block of login questions becomes a block of password questions, while symmetric noise would scatter them over all other topics.

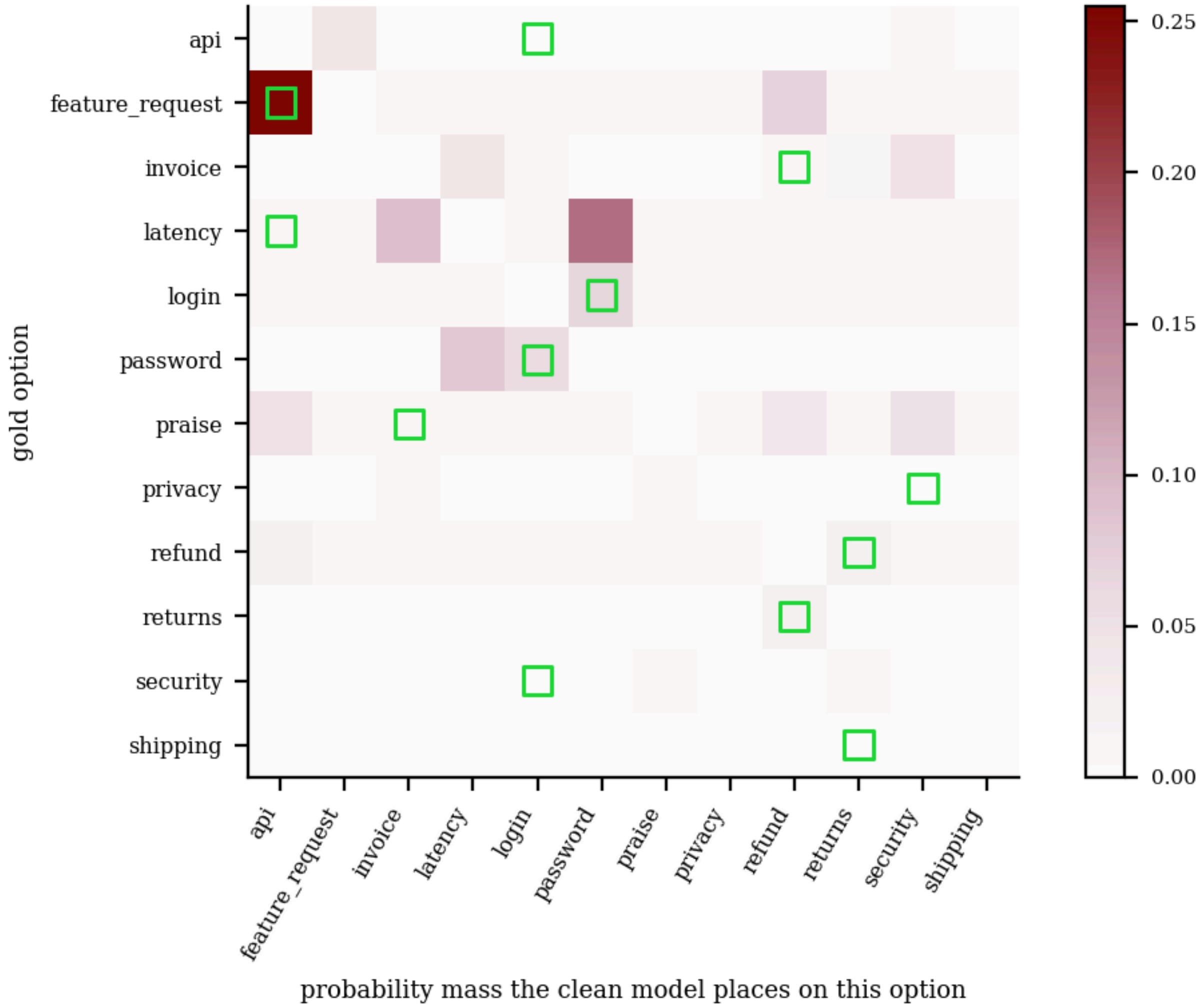

**Figure 13.** Where the confusable-pair noise sends a label. Shade: mean probability the clean full-data model puts on a wrong option (12-way topic task, 270 held-out items, diagonal removed); orange squares: the single option each gold label is flipped to.

### 9.1 Accuracy

The reference is the clean full-pool model (seed 0, 86.9%) and, for judging whether a difference matters, the four clean seeds (86.9% with s.d. 0.08). Every row of Table 10 is a single fine-tune on the full 2,818-row pool in which a share $p$ of the gold labels was replaced. The runs give: symmetric 10%: 86.7% (-0.2 points, McNemar p=0.909); symmetric 20%: 83.5% (-3.4 points, McNemar p<0.001); symmetric 40%: 78.1% (-8.7 points, McNemar p<0.001); confusable-pair 20%: 79.4% (-7.4 points, McNemar p<0.001). A line through the origin fitted to the available runs loses 2.0 points per 10% of symmetric corruption and 3.7 points per 10% of confusable-pair corruption; it is a summary, and the 10% symmetric run lies well below it. The 10% symmetric run is the only one without a detectable cost, and that is a statement about one run. Its paired 95% interval is -1.7 to +1.4 points around -0.2, so it rules out a cost above about 1.7 points but not a cost of one point. It also changes 8.1% of decisions against 9.5% between two clean seeds.

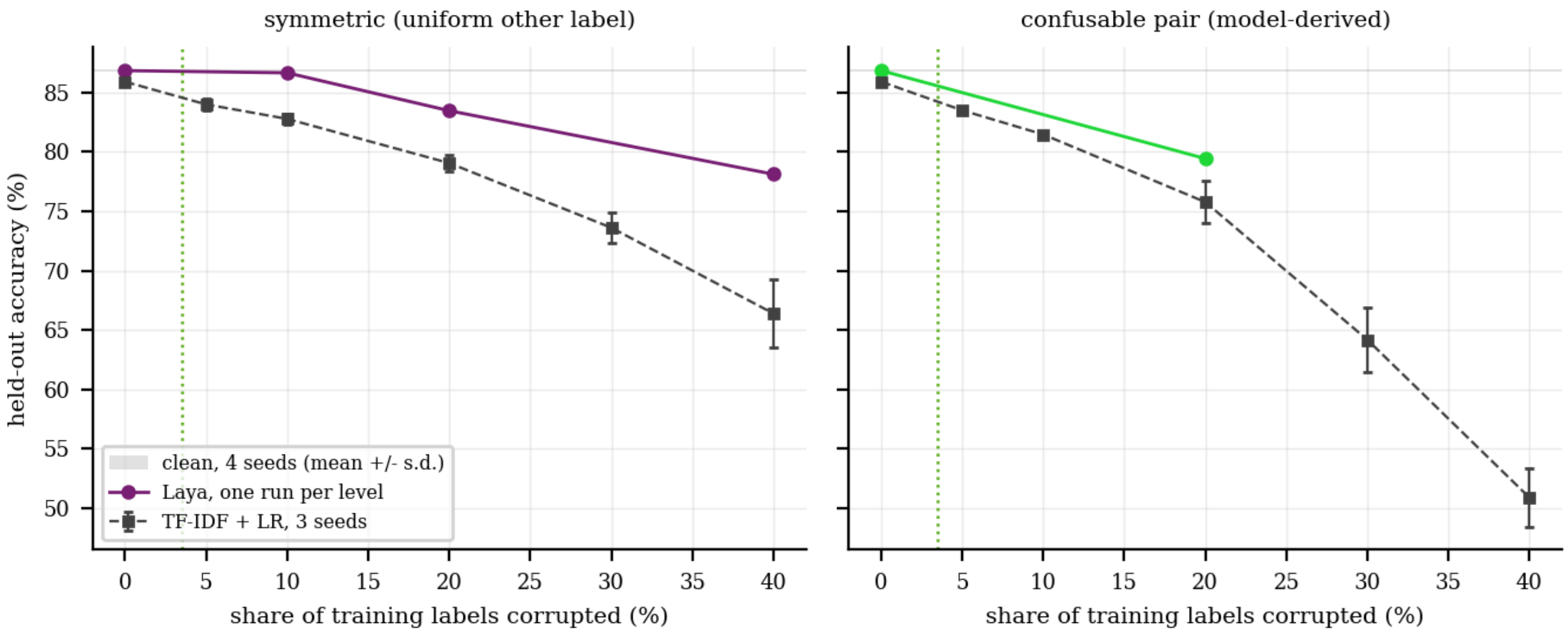

**Figure 14.** Held-out accuracy against label corruption on the full 2,818-row pool. The red dotted line marks the disagreement rate (3.5%) between the gold labels and an independent LLM labeller on the held-out items. TF-IDF + LR is trained on the very same corrupted labels.

**Table 10.** Label noise on the full pool (one run per cell). "rows flipped" is out of 2,818; Δ is against the clean seed-0 model on identical items with the exact McNemar p-value, and is marked when it lies inside two seed s.d. of the clean runs; over-confidence is mean confidence minus accuracy (points); flips is the share of the 1,065 decisions that differ from the clean seed-0 model (two clean seeds differ on the value in the first row); broke/healed split the flips by whether the clean model was right; the last column is TF-IDF + LR (3 seeds) on the same corrupted labels.

| Noise | p | rows flipped | accuracy [95% CI] | Δ vs clean (McNemar) | ECE | over-confidence | flips % | broke / healed % | TF-IDF acc. |
|---|---|---|---|---|---|---|---|---|---|
| clean | 0% | 0 | 86.9 ± 0.1 (4 seeds) | n/a | 7.0 | 6.6 | 9.5 | n/a | 85.9 |
| sym | 10% | 263 | 86.7% [84.5-88.6] | -0.2 (p=0.909) | 6.4 | -4.6 | 8.1 | 3.7 / 3.5 | 82.8 |
| sym | 20% | 548 | 83.5% [81.1-85.6] | -3.4 (p<0.001) | 13.1 | -12.3 | 10.9 | 6.4 / 3.0 | 79.1 |
| sym | 40% | 1103 | 78.1% [75.5-80.5] | -8.7 (p<0.001) | 24.4 | -24.3 | 17.7 | 11.9 / 3.2 | 66.4 |
| conf | 20% | 548 | 79.4% [76.9-81.8] | -7.4 (p<0.001) | 8.9 | -7.2 | 13.9 | 10.0 / 2.6 | 75.8 |

The TF-IDF classifier, trained on the identical corrupted labels (three seeds), changes accuracy by -6.9 points at 20% symmetric noise and by -10.1 points at 20% confusable-pair noise, against -3.4 and -7.4 points for Laya. At 40% symmetric noise the changes are -19.5 points for TF-IDF and -8.7 for Laya. TF-IDF changes by -35.1 points at 40% confusable-pair noise, for which no Laya fine-tune exists (Figure 14, dashed line). Only the levels listed in Table 10 were fine-tuned for Laya; the TF-IDF classifier, being cheap, was run at every level.

## 9.2 Calibration

Figure 15 shows ECE, mean confidence and over-confidence under noise. The clean models have an ECE of 7.0% (grey band: mean ± s.d. over four seeds). Every noisy run states lower probabilities than the clean model and falls below its own accuracy. Mean confidence minus accuracy is -4.6 (symmetric 10%), -12.3 (symmetric 20%), -24.3 (symmetric 40%), -7.2 (confusable-pair 20%) points, against +6.6 for the clean models. The model therefore moves from over-confident to under-confident, in each of the 4 runs (one seed each). ECE does not rise with it at every level, because the gap changes sign and passes through zero. It is 6.4% (symmetric 10%), 13.1% (symmetric 20%), 24.4% (symmetric 40%), 8.9% (confusable-pair 20%) against 7.0% clean, so at 10% symmetric and at 20% confusable-pair noise the ECE is close to the clean value while the model is clearly under-confident. Two mechanisms are mixed in this result and no run separates them. The wrong targets spread the cross-entropy over the wrong option, and the trainer fits its temperatures on a calibration slice from the same corrupted file, so the slice itself carries wrong labels.

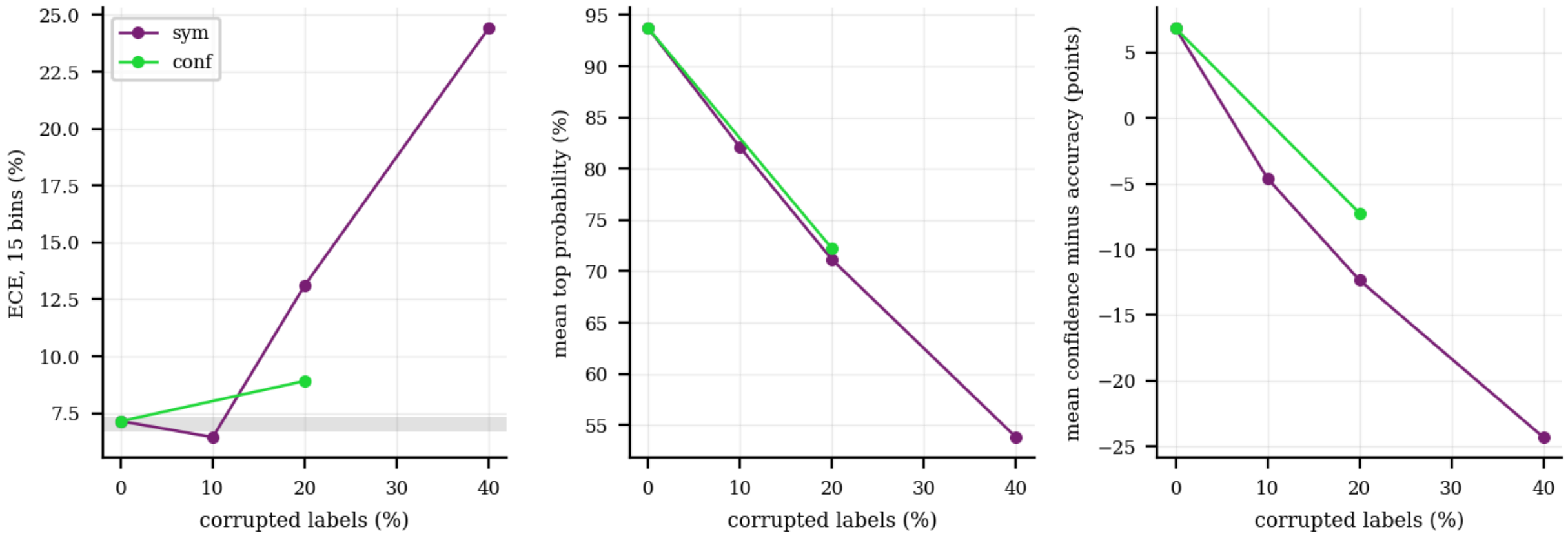

**Figure 15.** Calibration under label noise (grey band in the left panel: clean full-data models, mean ± s.d. of 4 seeds). Noise lowers the stated probability faster than it lowers accuracy, so in every noisy run the mean confidence ends below the accuracy (right panel: negative gap), where the clean models are over-confident.

## 9.3 Which decisions change

Accuracy is a net number, so Figure 16 splits the decisions that change relative to the clean seed-0 model into those that were right and became wrong (broke) and those that were wrong and became right (healed). The dashed line is the churn between two *clean* seeds, 9.5% of decisions on average over 6 pairs: a noisy model that differs from the clean one by less than that has not changed more than a re-seeding would. Figure 17 asks which of the correct decisions are the fragile ones: the items in the lowest confidence quartile of the clean model are the first to be lost.

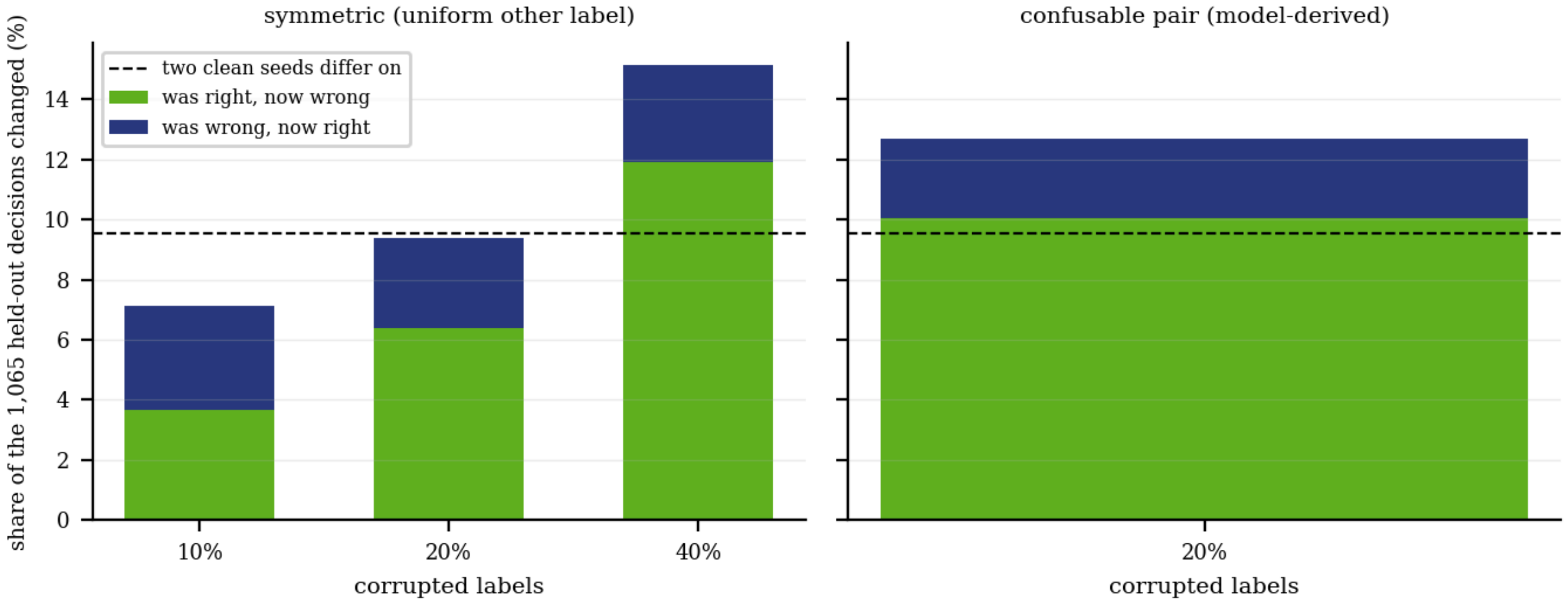

**Figure 16.** Decisions that change relative to the clean seed-0 model. Dashed line: two clean full-data seeds already disagree on 9.5% of decisions (mean of 6 pairs), so only the part of a bar above that line is attributable to the noise.

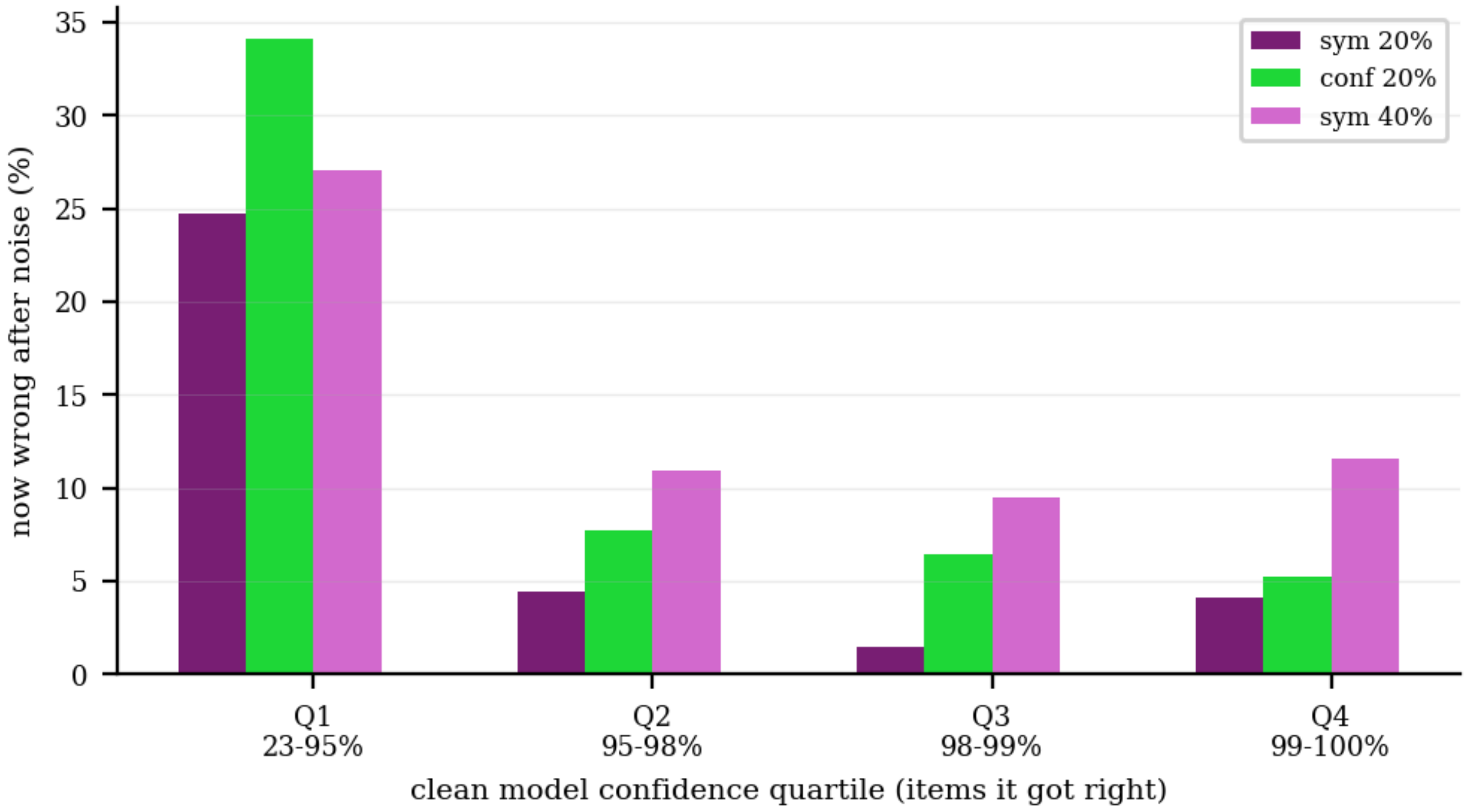

**Figure 17.** Which correct decisions the noise breaks. Items the clean model got right are split into quartiles of its own confidence; the bars give the share of each quartile that the noisy-label model answers wrongly.

### 9.4 Which tasks suffer

Figure 18 gives the change in task accuracy at each noise level. With 42 to 346 held-out items per task a change of less than about five points on the smaller tasks is within what a different seed would produce.

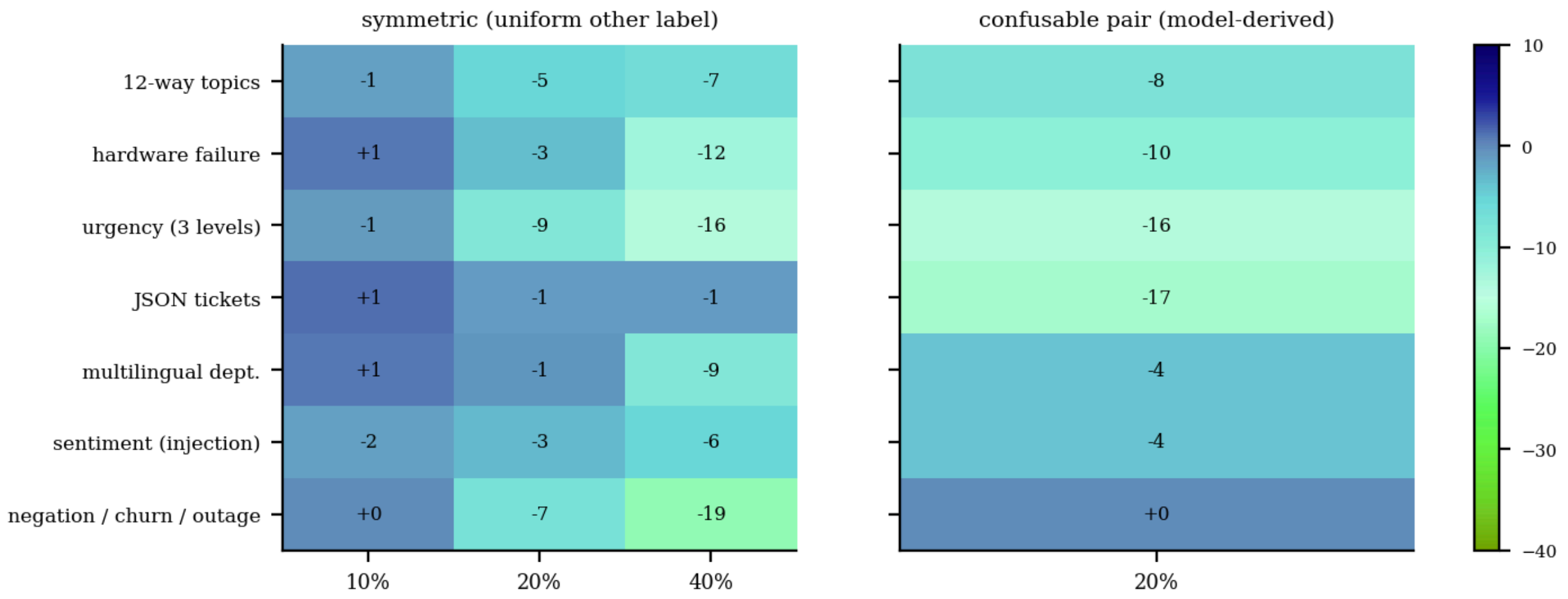

**Figure 18.** Change in task accuracy (points) relative to the clean seed-0 model, per noise level. Tasks with few held-out items (minor, urgency) move in steps of several points per item and should be read with their Wilson intervals in the table.

## 10 Noisy rows or fewer rows?

Corruption may cost more when there are few rows, and that matters for budgeting: a practitioner who can afford only a few hundred rows is also the one least able to throw away the doubtful ones. Table 11 and Figure 19 repeat the 20% corruption at 50% of the pool (symmetric only) and at the full pool (both kinds). Each noisy run is compared with the clean run on the same rows, and its accuracy is translated into an equivalent number of clean rows by inverting the fitted pooled curve.

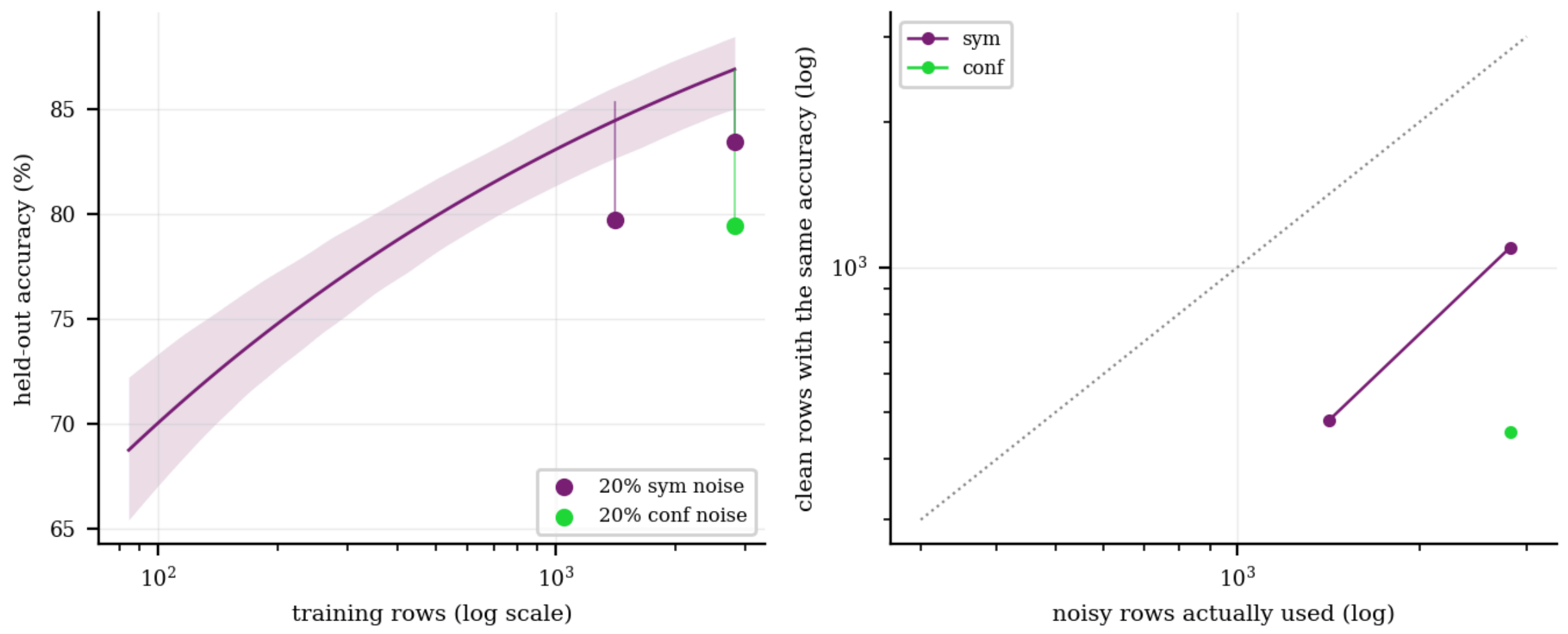

**Figure 19.** Noise at equal size. Left: runs with 20% corrupted labels (points) against the clean curve; vertical segments join each run to the clean run on the same rows. Right: how many clean rows give the same accuracy on the fitted curve, against the number of noisy rows actually used (dotted: equal).

**Table 11.** Twenty percent of the labels corrupted, at 50% and 100% of the pool. "Equivalent clean rows" is the number of clean rows at which the fitted pooled curve has the accuracy of the noisy run; the ratio is that number divided by the rows actually used.

| Noise | rows | seed | clean run, same rows | 20% corrupted | Δ (McNemar) | equivalent clean rows | ratio |
|---|---|---|---|---|---|---|---|
| sym | 1,413 | 0 | 85.4% [83.1-87.3] | 79.7% [77.2-82.0] | -5.6 (p<0.001) | 481 | 0.34 |
| sym | 2,818 | 0 | 86.9% [84.7-88.8] | 83.5% [81.1-85.6] | -3.4 (p<0.001) | 1,098 | 0.39 |
| conf | 2,818 | 0 | 86.9% [84.7-88.8] | 79.4% [76.9-81.8] | -7.4 (p<0.001) | 455 | 0.16 |

## 11 The independent labeller as a noise meter

The held-out gold labels come from the same generator as the training labels, so they carry the same kind of error. An independent labeller (a model from a different family, shown only the message and the options) was run on all 1,065 held-out items and gives the same answer as the gold label on 96.5% of them, 37 items differing (Table 12, Figure 20). This number works as a ruler for the noise sweeps because the disagreements are not uniform and because they fall where the model errs.

**Table 12.** Agreement between the gold labels and an independent labeller, and Laya's accuracy (mean of the four full-pool seeds) scored against each of the two labellings.

| Set | items | labeller agrees with gold (%) | Laya vs gold (%) | Laya vs labeller (%) |
|---|---|---|---|---|
| probe (mixed) | 224 | 97.3 | 90.2 | 89.6 |
| sentiment (injection) | 106 | 99.1 | 95.3 | 94.3 |
| JSON tickets | 58 | 96.6 | 90.9 | 90.9 |
| hardware failure | 82 | 95.1 | 94.2 | 92.7 |
| 12-way topics | 195 | 94.4 | 92.4 | 91.3 |
| multilingual | 322 | 97.2 | 74.7 | 73.6 |
| urgency | 78 | 94.9 | 91.3 | 90.1 |

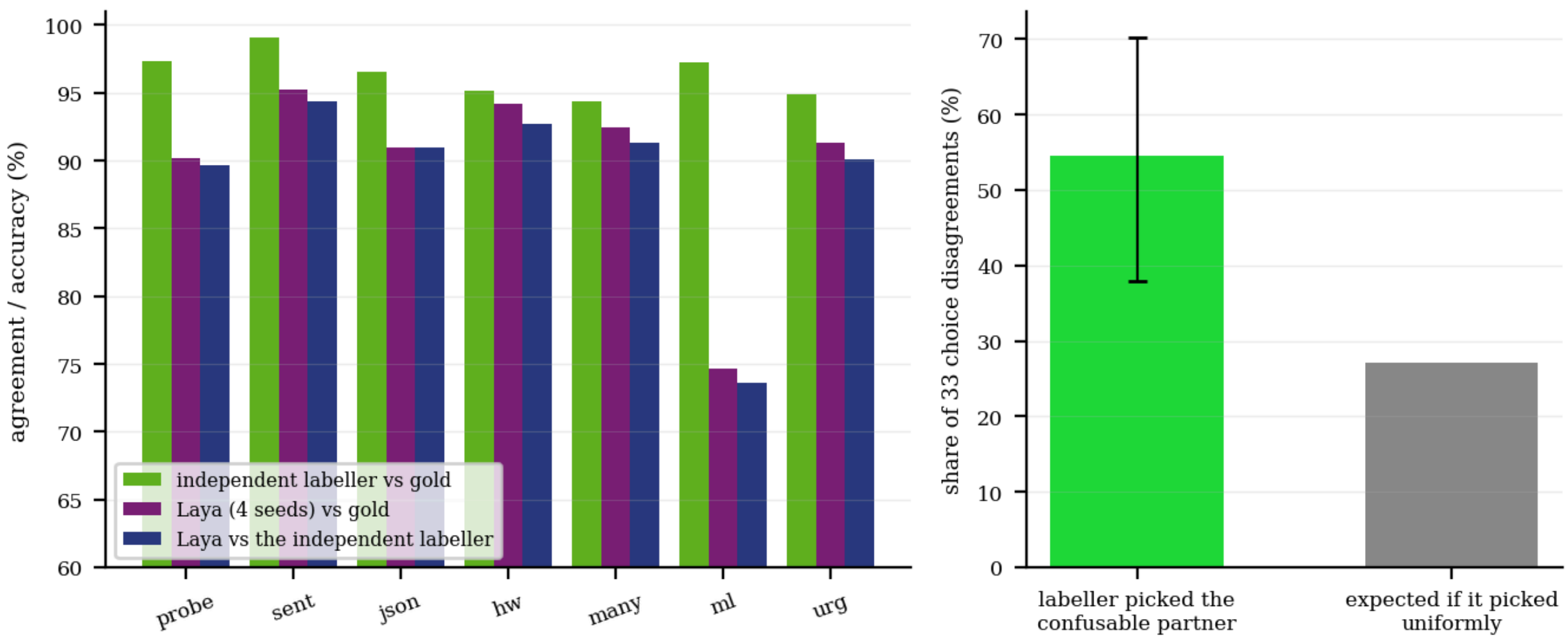

**Figure 20.** Left: per held-out set, agreement of an independent labeller with the gold labels, and Laya's accuracy against each labelling. Right: when the independent labeller disagrees on a choice question, how often it names the same partner label that the confusable-pair noise uses (partner map read from the same items, so this count is the optimistic one; the text gives leave-one-out counts).

On choice questions the labeller disagreed 33 times, and in 18 of those (55%, Wilson interval 38-70%) it named exactly the partner option that the confusable-pair noise uses, against 27% expected if it had chosen among the wrong options uniformly. The partner map was read from the same held-out items, these ones included, so that count flatters the match. Leaving each item out of the map that scores it gives 16 of 33 (48%, Wilson interval 33-65%), and leaving out all 33 contested items gives 11 (33%, 20-50%). The labeller's disagreements therefore lean towards the confusable partner. How far depends on whether the map may see the contested items, and with 33 items the data support no more than "somewhat closer to the confusable-pair model than to the symmetric one". A large share of the disagreements fall elsewhere. Figure 13 draws the structure that the partner rule exploits.

Laya (four full-pool seeds) is wrong on 39.9% of the 37 items on which the two labellers differ and on 12.2% of the others, so 15 of its 140 errors on average sit on contested items. If a prediction were counted correct when it matches *either* labeller, accuracy would be 88.0% against 86.9% as scored. Label disagreement is thus a real but small part of the error: it explains 11% of the mistakes, and the remaining 89% are items on which both labellers agree and the model is wrong. Reading the noise sweeps of Section 9 at that rate (linear interpolation between the clean model and the nearest noisy runs) puts the accuracy cost of 3.5% symmetric noise at -0.1 points and of 3.5% confusable-pair noise at -1.3 points. The larger of the two is larger than twice the standard deviation of the clean seeds. Both are interpolations that assume a straight line between single runs; scaling the paired 95% intervals of the 10% symmetric and the 20% confusable-pair run to 3.5% gives -0.6 to +0.5 and -1.6 to -0.9 points.

## 12 A practical budget

Table 13 turns the curves into pool sizes for a target accuracy, and Table 8 (Section 7) gives the same for each task. On the clean seed-mean curve, 82% pooled accuracy needs about 859 rows (24 per question-option cell) and 85% about 1,456 (40 per cell). The TF-IDF classifier needs 1,451 and 2,099. With 20% of the labels wrong at random, the full pool reaches 83.5%, which on the fitted clean curve is worth about 1,098 clean rows. A noisy pool of 2,818 rows buys what 39% as many clean rows would. This rests on one run and an inversion of the fitted curve, so only the order of magnitude is reliable.

The per-task budgets differ a great deal. To come within five points of its full-pool accuracy a task needs as few as 2 rows per cell (negation / churn / outage) or as many as 66 (multilingual dept.). Within two points the range is 7 to 82 rows per cell. These budgets are read off a coarse grid and many sizes have one seed. The seed spread at full size is already 0.6 to 2.3 points per task, so a "within two points" entry is one grid step wide at best. For a pooled accuracy near 82% on this kind of synthetic task, the data support a working rule: collect on the order of 24 rows per question-option cell, and add rows only for the tasks whose curve is still rising.

**Table 13.** Rows of the whole pool needed to reach a pooled accuracy, with rows per question-option cell in brackets (the pool has 36 cells). Laya and TF-IDF use the seed-mean clean curves (log-linear interpolation between measured sizes); the last two columns use the sizes at which a run with 20% corrupted labels exists (50 and 100% for symmetric noise, so a line through two points; 100% only for confusable-pair noise, so "not run"), and read "not reached" where the noisy curve does not get that high inside the pool.

| target accuracy | Laya, clean labels | TF-IDF + LR, clean labels | Laya, 20% symmetric noise | Laya, 20% confusable-pair noise |
|---|---|---|---|---|
| 78.0% | 442 (12) | 1,006 (28) | 1,413 (39) | not run |
| 82.0% | 859 (24) | 1,451 (40) | 2,149 (60) | not run |
| 84.0% | 1,226 (34) | 1,856 (52) | not reached | not run |
| 85.0% | 1,456 (40) | 2,099 (58) | not reached | not run |
| 85.9% | 1,714 (48) | 2,804 (78) | not reached | not run |

## 13 Limits and threats to validity

Training rows and held-out items were produced by one LLM and screened by a second pass of the same LLM, so every label is a model opinion and accuracy here is agreement with that labeller. An independent labeller from another family agrees with the held-out gold labels on 96.5% of the 1,065 items. At least 3.5% of the held-out labels are therefore contestable, and accuracies within that margin of the ceiling cannot be told apart from a perfect model. Nothing in this paper is evidence about human labels: human annotators make different, instance-dependent errors that neither of our noise models imitates.

The data are synthetic and in-distribution. Generator, screening, training pool and held-out sets are one pipeline (Section 2.2). Curves measured on a held-out set from the same generator are optimistic about real traffic, and probably steeper than a curve measured on real messages would be, because generated messages for one answer resemble each other. The rows-per-option budgets are therefore lower bounds for any real deployment.

Every run fine-tunes the same ModernBERT-large checkpoint with the same trainer and hyper-parameters. We changed neither the learning rate nor the loss when the data shrank, although the optimum surely moves. The equal-update control shows how much that matters, but only for one setting.

Seed counts are low. The clean curve has seeds per size of 5%: 1, 10%: 1, 20%: 2, 35%: 1, 50%: 2, 75%: 1, 100%: 4; noise runs have one seed, and one extra seed was run only for 20% and 50% of the data without noise. Seed-to-seed standard deviations are therefore estimated from two to four values, and a single noise run can differ from its twin by more than the effect it is meant to show. Wherever a difference is below twice the clean-seed standard deviation, the tables say so. That standard deviation comes from the four full-pool seeds, which share all their rows and are the quietest case; the seed spread at small sizes is larger (Table 4). The 100% point is the earlier full-data fine-tune with four seeds. It was trained by the same trainer from disk-stored weights, while the new runs keep weights in memory. The evaluation code is the same. Two clean full-data seeds already disagree on 9.5% of held-out decisions.

The fit has seven fractions and three parameters. The plateau and rows-to-95% are extrapolations whenever they lie beyond the pool, and their bootstrap intervals only cover held-out sampling and seed sampling, not the choice of functional form. The measured "within 2 or 5 points" columns in the budget table do not depend on the form.

7 runs are compared with a clean reference run on paired items (some of them in two tables), and the p-values are exact and uncorrected. A p-value between 0.01 and 0.05 in a table of this size should be treated as suggestive. A p-value above 0.1 from a single run is not evidence of no effect: the 10% symmetric run has a paired 95% interval of -1.7 to +1.4 points around -0.2.

The confusable partner of each option was read from the clean models' behaviour on the held-out items, so the noise design knows which options those models confuse on this distribution. No trained model saw a held-out item, but the confusable-pair sweep is not blind, and the same items define the partner map and the labeller comparison in Section 11 (which the leave-one-out count there addresses only in part). The calibration slice of a noisy run is cut from the corrupted file. The under-confidence of the noisy models therefore mixes the effect of wrong targets in the loss with temperatures fitted on wrong labels. 15 of the 35 configurations in the manifest were run; the others were dropped for lack of time and appear in no figure or table.

Training shared a GPU with unrelated jobs, and a few runs were restarted after device errors. A restarted run is a fresh run with the same seed. The fine-tune is not bit-reproducible on shared hardware, which is part of the seed variance reported here.

## 14 References and related work

The title and abstract of every entry below were read on the Hugging Face papers index or on arXiv on 10 November 2026 before it was cited. The search for related work was by title and abstract only and is not exhaustive.

Kaplan et al. [1] showed that language-model loss falls as a power law in data, model size and compute, with fits that hold over many orders of magnitude. Our setting differs in what matters for extrapolation. We fine-tune a classifier and measure accuracy, which has a ceiling set by the test labels. Our data range is about 20 to 1, where Kaplan et al. cover seven orders of magnitude. We borrow the habit of fitting a curve and reporting where it flattens, and make no claim that the form is universal. Hernandez et al. [9] measured in a fine-tuning setting that pre-training behaves like a multiplier on the fine-tuning data set, with a power law in its size at small sizes. That is the regime of this paper, but their quantity is loss on language data and ours is accuracy on a classification task.

The noisy-label survey [3] organises the field by noise type (instance-independent symmetric and class-conditional noise, instance-dependent noise) and by remedy. Confident learning [2] estimates the class-conditional transition matrix from model probabilities. We use the same ingredient, the probabilities of clean models, to *generate* the confusable-pair noise, whereas confident learning uses them to *find* noise. Zhang et al. [4] showed that large networks can fit arbitrary labels, and Rolnick et al. [5] that they can still generalise when correct labels are outnumbered, at the price of more data and also when the wrong labels are biased towards confusing classes. Both concern image benchmarks and training from scratch, whereas we fine-tune a pre-trained encoder for about 160 updates on under 3,000 rows. We did not measure training-set fit, so this paper cannot say whether the encoder memorised the corrupted labels. Rolnick et al. also report that the dataset-size cost of noise is related to how heavily the correct labels are diluted, and our exchange rate between corrupted and clean rows (Section 10) examines that at a much smaller scale. For fine-tuned language models, Li and Sen [10] compare label, grammatical and typographical noise on three model families and three tasks and find that label noise degrades performance most. They do not vary the structure of the label noise or relate it to an annotator's disagreement rate.

Ye et al. [11] study classifiers trained on labels that an LLM wrote zero-shot or few-shot, and argue that such noise is understudied relative to synthetic and human noise. They propose a method (SiDyP) that refines noisy labels through neighbourhood label distributions in embedding space, and report average gains of about 7 points for a BERT classifier. Merdjanovska et al. [12] show for named entity recognition that fine-tuning ignores LLM annotation noise and that error types (missing mentions against type errors) need different loss reweighting, with gains of 0.8 to 2.0 points at 15 to 40% noise. NoiseBench [13] shows, again for entity recognition, that real noise (expert, crowd, automatic and LLM errors) is considerably harder than simulated noise and that robust-learning methods fall short of their upper bound. These works motivate our two noise structures and our caution that injected noise is a stand-in. We study no remedy and no entity recognition. We ask how much injected noise a closed-set decision encoder tolerates, and we compare it with a measured disagreement between two LLM labellers.

Rafe and Das [6] benchmark typed-decision checkpoints against supervised classifiers and generative models and report that a fine-tuned DeBERTa-v3-large is the most accurate on all but one of the labelled benchmarks when task labels exist. That raises the question of how many labels. TypedBench [7] argues that such models must be judged on calibration, wording sensitivity and induced decisions. It also reports calibration error against a finite-sample noise floor; we follow the spirit by quoting the binomial standard error beside differences (Sections 4 and 6). The Turkish model of [8] documents that temperature scaling can raise calibration error on unseen questions and flags the numbers that were shaped by reading test results, a caution we apply in Section 13.

### What is and is not new

Four results are not new: accuracy rises with data and saturates; class-conditional label noise hurts more than uniform noise; pre-trained encoders tolerate moderate noise; and stated confidence drifts when labels are wrong.

As far as we found, the new part is the combination for typed-decision encoders trained on LLM-labelled synthetic rows (data scaling and noisy LLM labels are each studied, see above). We measure these quantities for Laya's own trainer, with seed repeats and an equal-update control. We add per-task rows-per-option budgets with bootstrap intervals, a noise sweep set against the unit that practitioners can observe (the disagreement between two LLM labellers on the same items), and an exchange rate between corrupted rows and clean rows. We make no claim about human labels. Other backbones, other tasks and data from other pipelines are also out of scope.

Statistical tools (Wilson intervals, the McNemar test, the bootstrap) and the Laya model card are listed in the general reference list at the end.

## Appendix A Every run

**Table A-1.** Every fine-tune used in this paper. The four full-pool rows were trained earlier with the same recipe; the others come from the manifest in a10_sub.py (names: f<percent>_s<seed>[_<noise><percent>][_ep<epochs>]).

| run | labels | epochs | rows | flipped | updates | pooled acc. (%) | ECE (%) |
|---|---|---:|---:|---:|---:|---:|---:|
| f005_s0 | clean | 4 | 143 | 0 | 12 | 74.3 | 6.7 |
| f010_s0 | clean | 4 | 281 | 0 | 16 | 74.6 | 9.6 |
| f020_s0 | clean | 4 | 564 | 0 | 32 | 78.7 | 5.3 |
| f035_s0 | clean | 4 | 987 | 0 | 56 | 82.7 | 5.2 |
| f050_s0 | clean | 4 | 1413 | 0 | 80 | 85.4 | 5.9 |
| f075_s0 | clean | 4 | 2108 | 0 | 120 | 87.0 | 6.1 |
| f100_s0_sym20 | sym 20% | 4 | 2818 | 548 | 160 | 83.5 | 13.1 |
| f100_s0_conf20 | conf 20% | 4 | 2818 | 548 | 160 | 79.4 | 8.9 |
| f020_s1 | clean | 4 | 564 | 0 | 32 | 80.9 | 3.0 |
| f100_s0_sym40 | sym 40% | 4 | 2818 | 1103 | 160 | 78.1 | 24.4 |
| f005_s0_ep54 | clean | 54 | 143 | 0 | 162 | 76.2 | 20.6 |
| f020_s0_ep20 | clean | 20 | 564 | 0 | 160 | 80.6 | 13.1 |
| f050_s0_sym20 | sym 20% | 4 | 1413 | 275 | 80 | 79.7 | 7.4 |
| f100_s0_sym10 | sym 10% | 4 | 2818 | 263 | 160 | 86.7 | 6.4 |
| f050_s1 | clean | 4 | 1413 | 0 | 80 | 84.3 | 6.4 |
| full pool, seed 0 | clean | 4 | 2818 | 0 | 160 | 86.9 | 7.2 |
| full pool, seed 1 | clean | 4 | 2818 | 0 | 160 | 86.9 | 7.3 |
| full pool, seed 2 | clean | 4 | 2818 | 0 | 160 | 86.9 | 7.0 |
| full pool, seed 3 | clean | 4 | 2818 | 0 | 160 | 86.8 | 6.6 |

15 of the 35 configurations defined in the code were run (rows of Table A-1). The others were not run for lack of time and are not used anywhere in the paper.

## Appendix B Definitions and reproduction

**Accuracy** is the share of held-out items whose arg-max option equals the gold option. **ECE** is the 15-bin expected calibration error of the top probability. **Over-confidence** is mean top probability minus accuracy. **Flip** is an item whose arg-max option differs between two models; a flip is *broken* when the reference model was right and *healed* when it was wrong. **Seed s.d.** is the sample standard deviation of pooled accuracy over independently seeded runs on the same rows. **Binomial s.e.** is the standard error of one run's accuracy over the 1,065 items, sqrt(p(1−p)/n). **Retention** is task accuracy at a fraction divided by task accuracy at 100%. **Equivalent clean rows** is the row count at which the fitted pooled clean curve reaches the accuracy of a noisy run. **Option cell** is one (question schema, answer option) pair. **Partner** is the single wrong option that the clean full-data models give the most probability on, averaged over every held-out item whose gold answer is the option in question.

The code is in the study folder next to the models. `a10_sub.py` defines the nested subsets, the two noise processes and the manifest. `a10_run.py` is a sequential, resumable driver that waits for free GPU memory and disk. `a10_train.py` calls Laya's trainer in-process and scores the new model on the seven held-out sets without writing weights to disk, and `a10_tfidf.py` trains the bag-of-n-grams classifiers on the same subsets and corrupted labels. One result file per run is stored in `results/a10_<name>.json` with the per-item probabilities, so `lp_u10.py` (with `a10_ana.py`, `a10_core.py`, `a10_fig_scale.py`, `a10_fig_noise.py`, `a10_tab.py`) recomputes every number in this paper without loading any model. To rerun the study: `python a10_run.py`, then `python a10_tfidf.py`, then `python build_long.py a10`.

**Table A-2.** Fingerprints (first 16 hex digits of SHA-256) of the shared inputs.

| file | SHA-256 |
|---|---|
| results/a10_tfidf.json | 9936271f07fd73bf… |
| results/a10_partner.json | dba4feaf6a7f4ad5… |
| results/xlabel_gemma4-31b.json | 28f1fa32695a39f6… |

The bootstrap uses fixed seeds (resampling of items and seeds, 300 refits for the pooled curve and for each task; 400 resamples for the paired gaps). The saturating fit is a grid search over the exponent followed by a closed-form least-squares step for the other two parameters, so it has no random start.